\documentclass[preprint,12pt]{elsarticle}

\usepackage{graphicx}
\usepackage{amssymb}
\usepackage{hyperref}

\usepackage{lineno}

\biboptions{authoryear}

\journal{Cognition}

\begin{document}

\begin{frontmatter}

\title{Exploration and Exploitation of Victorian Science\\ in Darwin's Reading Notebooks}
\author[1,2]{Jaimie Murdock}
\ead{jammurdo@indiana.edu}
\author[1,3,4]{Colin Allen}
\ead{colallen@indiana.edu}
\author[1,2,5,6]{Simon DeDeo \corref{cor1}}
\ead{simon@santafe.edu}
\cortext[cor1]{To whom correspondence should be addressed.}
\address[1]{\footnotesize Program in Cognitive Science, Indiana University, Bloomington, IN 47405, USA}
\address[2]{\footnotesize School of Informatics and Computing, Indiana University, 919 E. 10th Street, Bloomington, IN 47408, USA}
\address[3]{\footnotesize Department of History and Philosophy of Science and Medicine, \\ Indiana University, Bloomington, IN 47405, USA}
\address[4]{\footnotesize School of Humanities and Social Sciences, Xi'an Jiaotong University, Xi'an, China}
\address[5]{\footnotesize Department of Social and Decision Sciences, Carnegie Mellon University, 5000 Forbes Avenue, BP 208, Pittsburgh, PA 15213, USA}
\address[6]{\footnotesize Santa Fe Institute, 1399 Hyde Park Road, Santa Fe, NM 87501, USA}




\begin{abstract}
Search in an environment with an uncertain distribution of resources involves a trade-off between exploitation of past discoveries and further exploration. This extends to information foraging, where a knowledge-seeker shifts between reading in depth and studying new domains. To study this decision-making process, we examine the reading choices made by one of the most celebrated scientists of the modern era: Charles Darwin. From the full-text of books listed in his chronologically-organized reading journals, we generate topic models to quantify his local (text-to-text) and global (text-to-past) reading decisions using Kullback-Liebler Divergence, a cognitively-validated, information-theoretic measure of relative surprise.  Rather than a pattern of surprise-minimization, corresponding to a pure exploitation strategy, Darwin's behavior shifts from early exploitation to later exploration, seeking unusually high levels of cognitive surprise relative to previous eras. These shifts, detected by an unsupervised Bayesian model, correlate with major intellectual epochs of his career as identified both by qualitative scholarship and Darwin's own self-commentary. Our methods allow us to compare his consumption of texts with their publication order. We find Darwin's consumption more exploratory than the culture's production, suggesting that underneath gradual societal changes are the explorations of individual synthesis and discovery. Our quantitative methods advance the study of cognitive search through a framework for testing interactions between individual and collective behavior and between short- and long-term consumption choices. This novel application of topic modeling to characterize individual reading complements widespread studies of collective scientific behavior.
\end{abstract}


\end{frontmatter}

\vspace{1cm}


\section{Introduction}
\label{S:1}

The general problem of ``information foraging'' \citep{Pirolli1999} in an environment about which agents have incomplete information has been explored in many fields, including cognitive psychology \citep{Todd2012,Hills2015}, neuroscience \citep{Cohen2007}, economics \citep{March1991,Azoulay-Schwartz2004}, finance \citep{Uotila2009}, ecology \citep{Stephens1986,Eliassen2007}, and computer science \citep{Sutton1998}. In all of these areas, the searcher aims to enhance future performance by surveying enough of existing knowledge to orient themselves in the information space.

Individual scientists and scholars can be viewed as conducting a cognitive search \citep{Todd2012} in which they must balance \emph{exploration} of ideas that are novel to them against \emph{exploitation} of knowledge in domains in which they are already expert \citep{Berger-Tal2014}. 
Researchers have studied the exploration-exploitation trade-off in cognitive search at timescales of minutes up to years and decades. Laboratory experiments on visual attention are one example of this balancing act on short timescales \citep{chun1996just}, while studies of the recombination of patented technologies demonstrate long-term group behavior \citep{Youn2014}. New advances in the digitization of historical archives enable longitudinal study of how individuals explore and synthesize the work of their contemporaries and predecessors over the course of a lifetime.

As one of the most successful and celebrated scientists of the modern era, Charles Darwin's scientific creativity has been the subject of numerous narrative and qualitative studies \citep{Gruber1974,Johnson2010,VanHulle2014}. In part, these studies are possible because Darwin left his biographers careful records of his intellectual and personal life. These include records of the books he read from 1837 to 1860, a critical period which culminated in the publication of \emph{The Origin of Species}. Table~\ref{table-timeline} summarizes key events in Darwin's life.

\begin{table}[t]
\begin{center}
\textbf{Major Events in Charles Darwin's Life (1809-1882)}
\begin{tabular}{rl}
\hline 12 Feb 1809 & Born in Shrewsbury, England \\
22 Oct 1825 & Matriculates at University of Edinburgh \\
15 Oct 1827 & Admitted to Christ's College, Cambridge \\
27 Dec 1831 & Departs England aboard the \emph{HMS Beagle} \\
2 Oct 1836 & Return to England aboard the \emph{HMS Beagle} \\ \hline
\textbf{July 1837} & \textbf{First entries in reading notebooks} \\
Aug 1839 & Publication of \emph{The Voyage of the Beagle} (1st edition) \\
May 1842 & Writes the 1st Essay on Species \\
4 July 1844 & Writes the 2nd Essay on Species \\
Aug 1845 & Publication of \emph{The Voyage of the Beagle} (2nd edition) \\
1 Oct 1846 & Begins barnacle project \\
19 Feb 1851 & Publishes first volume of barnacle work \\
9 Sep 1854 & Begins sorting notes on natural selection \\
14 May 1856 & Starts writing ``large work'' on species \\
24 Nov 1859 & Publication of \emph{The Origin of Species} (1st edition)\\
\textbf{13 May 1860} & \textbf{Last entry in reading notebooks} \\ \hline
24 Feb 1871 & Publication of \emph{The Descent of Man} \\
19 Feb 1872 & Publication of \emph{The Origin of Species} (6th and final edition) \\
21 Apr 1882 & Dies at Down House in Kent, England \\ \hline
\end{tabular}
\end{center}
\caption{{\bf Timeline}. Major events in Charles Darwin's life, including those marked on Fig.~\ref{fig:kl_over_time}. This paper focuses on the critical period of his work from 1837 to 1860, leading to the publication of \emph{The Origin of Species}. See \citet{berra2009charles} for an expanded chronology.}
\label{table-timeline}
\end{table}

This article presents the first quantitative analysis of an important scientist's reading diaries, tracking how Darwin navigated the exploration-exploitation trade-off in choosing what to read. We link Darwin's reading records with the full text of the original volumes. We then use probabilistic topic models \citep{Blei2003,Blei2012} to represent the original text of each book Darwin read as a mixture of topics. We use information theory to measure the surprise, or unpredictability, of the next book that Darwin chose to read, compared to his past history of reading.

We present three key findings: 
\begin{enumerate}
\item Darwin's reading patterns switch between both exploitation and exploration throughout his career. This is in contrast to a pure surprise-minimization strategy that consistently exploits content within a local region before moving on. The general trend, as Darwin's career develops, is towards increasing exploration.
\item In comparison to the publication order of the texts Darwin read, Darwin's reading order shows higher average surprise. This indicates that the order in which the books were written by the scientific community is less surprising than the order in which Darwin read them.
\item Darwin's strategies fall into three long-term epochs, or behavioral modes characterized by distinct patterns of surprise-seeking. These epochs correspond to three biographically significant periods: Darwin's post-\emph{Beagle} studies, his extensive work on barnacles, and a final period leading to his synthesis of natural selection in the \emph{Origin of Species}.
\end{enumerate}

While the bulk of empirical studies in cognitive science are concerned with measuring population-level effects due to experimental manipulations, case studies play an important role in driving cognitive theorizing, experimentation, and modeling. For example, the case of the memory-impaired patient H.M. has driven many advances in cognitive neuroscience and computational models of memory (reviewed in \citet{SquireWixted2011}).  Other case studies, such as that of the frontal-lobe injury in Phineas Gage, provide important contrasts for later studies (reviewed in \citet{Macmillan2000}). 

Detailed longitudinal investigations of a single individual may involve repeated trials and changing strategies that cannot be observed in laboratory experiments involving a single task. Cognitive science could be enriched by using longitudinal studies such as ours to design laboratory studies with higher ecological validity. However, it is a challenging task to design laboratory studies of (for example) reading choices that take into account a subject’s extensive prior history of reading decisions.

An advantage of taking Charles Darwin as a case study is the the extensive attention that he has received from historians and biographers since his death. These studies provide a novel means for validating our mathematical tools. We will present a number of theoretical and empirical reasons why our methods measure cognitively-relevant features of a reader's experience. At the same time, the fact that our results also recover key features of Darwin's intellectual life provides additional support for the reliability of our methods.

We cannot claim Darwin's information foraging behavior is typical for either scientists or the population-at-large; however, our analytic methods may be applied to other case studies to find population-level generalizations, provided there is access to the full text of each reading.

Our approach contrasts with previous uses of topic modeling to analyze the large-scale structure of scientific disciplines \citep{Griffiths2004,Hall2008,Blei2007,Cohen2015} and the humanities \citep{Poetics41, JDH, Jockers2013}, which are each created through the collective effects of individual-level behavior. 
Previous models of historical records have focused on language use as an indication of larger shifts in style \citep{Hughes2012,Underwood2012}, learnability \citep{hills2015recent}, or content \citep{Michel2011,Goldstone2014,Klingenstein2014} of significant portions of publications in a field, including a study of \emph{Cognition} itself \citep{Cohen2015}. 

These works model the collective state of all published works at a particular date, but obscure the role of individual foraging behavior. By focusing on a single individual for whom ample records exist, we gain access to what \citet{Tria2014} describe as ``the interplay between individual and collective phenomena where innovation takes place''.

\section{Materials and Methods}
\subsection{Darwin's Reading Records}
Darwin was a meticulous record-keeper---starting in April 1838, he kept a notebook of ``books to be read'' and ``books read''. These records span the 23 years from 1837 to 1860, tracking his reading choices from just after his return to England aboard the \emph{HMS Beagle} to just after the publication of \emph{The Origin of Species}. We located the full-text of 665 of the 687 (96.7\%) English non-fiction mentioned in these reading notebooks through a variety of online digital libraries. See \ref{S1_Text} for details of corpus curation.

\subsection{Probabilistic Topic Models}
We model these texts using using Latent Dirichlet Allocation (LDA; \citet{Blei2003,Blei2012}), a type of probabilistic topic model. LDA is a generative model that represents each document as a bag of words generated by a mixture of topics. LDA topic models have been extensively used to analyze human-generated text across many domains \citep{Griffiths2004,Blei2007,Poetics41,JDH,Jockers2013}. Furthermore, topic models predict the behavior of human subjects in a variety of word association and disambiguation tasks \citep{Griffiths2007}. Topic models allow us to describe Darwin's reading as taking place in a $(k-1)$-dimensional space, the simplex, where a particular volume is described as a probability distribution, $\vec{p}$, over $k$ topics. To test the robustness of our results, we vary the number of topics, $k$. In the main body of the paper, we report for $k=80$ because we found that our results were robust to alternative values of $k$ and to differing random seeds for each; see \ref{S5_Text} for $k=\{20,40,60\}$.  Darwin's ``semantic voyage'' is the path he takes through this space from text to text.

\subsection{Cognitive Surprise and Kullback-Leibler Divergence}

We use probabilistic topic models to quantify the structure of the texts. Characterizing Darwin's reading choices then becomes the problem of comparing distributions over topics. Our goal is to quantify surprise: the extent to which a new reading either satisfies, or violates, the expectations a reader would have based on what came before. A high-surprise reading indicates that the reader has chosen a book that contains a new mixture of topics compared to what came before.

In our model, $\vec{p}$ and $\vec{q}$ are topic distributions over texts or an aggregation of texts. The reader is modeled in an information-theoretic fashion as building efficient mental representations of these distributions. Our task is then to quantify his exploratory behavior as he moves from one distribution to another.

To compare two distributions in this way we use the Kullback-Leibler (KL) divergence, first introduced by \citet{Kullback1951}. It is defined by
\begin{equation}
D_\mathrm{KL}(\vec{q} | \vec{p}) = \sum_{i=1}^k q_i \log_2 \frac{q_i}{p_i},
\end{equation}
where $\vec{p}$ is the distribution over topics that the reader has encountered before, and $\vec{q}$ the new distribution of topics that the reader encounters next. (In this paper, we consider two choices for $\vec{p}$: 1) the distribution over topics for the just previous book, and 2) the average over all books in the reader's past. Meanwhile, $\vec{q}$ will always refer to a distribution over topics for the next book a reader encounters.)

KL divergence can be thought of as a measure of surprise associated with a change in representation of a learner who encounters the unexpected in ways that require her to alter the way she represents the world. An agent expecting observations to be draw from probability distribution $\vec{p}$ will have those expectations violated if it arrives according to a different distribution $\vec{q}$. KL divergence quantifies the surprise in a particularly natural fashion; to aid the reader unfamiliar with this quantity, we present two ways to understand the mathematical structure of KL in \ref{SKL_Text}: first in terms of ``coding failure'' and second in terms of the rate of accumulation of Bayesian evidence against one distribution and in favor of the other.

The use of KL to measure cognitive surprise has had a number of recent successes. Vision researchers have used KL to track surprise in a visual scene: the places on the screen where new events most violate the viewer's previous assumptions; KL then accurately captures attention attractors in visual search tasks \citep{Demberg2008193,itti2009bayesian}. More generally, KL has seen wide use in the cognitive sciences; \citet{resnik1993selection}, for example, proposed the KL divergence as a measure of selectional preferences in language (reviewed in \citet{light2002statistical}). It has found use in many successful models of linguistic discrimination, including syntactic comprehension \citep{hale2001probabilistic,Levy20081126}, speech recognition \citep{COGS:COGS12167,COGS:COGS1267} and word sense disambiguation \citep{Resnik1997}. 

\subsection{Text-to-text and Text-to-past Surprise}

We use KL in two distinct ways. We measure the text-to-text surprise: given a distribution over topics for the text Darwin just read, how surprised is he upon encountering the next volume's topic distribution? Text-to-text surprise is a \emph{local} measure. We also measure the text-to-past surprise: given all of the volumes that Darwin has encountered so far, how surprised is Darwin by the text that comes next? Text-to-past surprise is a \emph{global} measure. 

If we define $\theta_i$ as the topic distribution for document $i$, then
text-to-text ($T2T$) and text-to-past ($T2P$) surprise are defined as
\begin{equation}
T2T(i) = D_{KL}(\theta_{i}|\theta_{i-1}),
\end{equation}
\begin{equation}
T2P(i) = D_{KL}\left(\theta_i\left| \frac{\sum_{j=0}^{i-1} \theta_j}{i}\right.\right).
\end{equation}

Text-to-text surprise and text-to-past surprise provide complementary windows onto Darwin's decision-making. Local decision-making, meaning the choice of the next text to read given the current one, is captured by text-to-text surprise. Global decision-making, the choice of which text to read given the entire history of reading to date, is captured by text-to-past surprise. Low surprise, in either case, is a signal of \emph{exploitation}, while high surprise indicates larger jumps to lesser-known topics, and thus of \emph{exploration}. These measures can be easily generalized to arbitrary text-to-$N$ surprise measures, representing the choice of the next reading given the history of readings within the past $N$ volumes or time periods. 

We characterize Darwin's decision process by the combination of text-to-text and text-to-past surprise. Exploration, indicated by high surprise, happens when a searcher is moving across a space not previously explored. Exploitation, indicated by low surprise, happens when a searcher has a sustained focus on material they are already familiar with.

These local and global behaviors do not have to align. For example, text-to-text surprise may be high (local exploration) at the same time that text-to-past surprise is low (global exploitation). This can happen if Darwin's readings interleave different topics that he has already seen. If, for example, Darwin alternates between readings in philosophy with readings in travel narratives, then each local jump will have high KL (a travel narrative is dominated by topics that are rare in a philosophical text, and vice-versa) and Darwin's readers will appear as a local exploration. However, once this pattern of alternation has been established, the average over past texts will include both philosophical and travel narrative topics, lowering the text-to-past surprise, and driving the system back towards global exploitation.
  
Conversely, text-to-text surprise can also be low while text-to-past surprise is high. This can happen if Darwin has recently begun a novel, but focused, investigation. In this situation, he focuses on a particular subset of topics that are under-represented in his overall history. If Darwin begins by focusing on philosophical texts, and then switches to travel narratives, his second (and subsequent) travel narrative readings will have low text-to-text surprise; but the average over past texts will be dominated by a long history of philosophical readings, leaving the text-to-past surprise high until he has accumulated so many readings of the latter type that they dominate the past average.

\subsection{Cultural Production and Null Reading Models}

Our methods also allow us to understand how Darwin selectively re-orders the products of his culture. To do this, we measure text-to-text and text-to-past surprise using the publication order of the texts that Darwin read. We can then ask questions about the extent to which Darwin's encounters with the products of his culture were more or less exploratory than the order in which they were produced. Does Darwin's reading order reduce the surprise relative to the publication order, or does it increase it?

All results are relative to a null reading model that holds Darwin's original reading dates fixed and re-samples without replacement from his original reading list. The title selection at each reading date is constrained to those titles published before that date. In contrast to a null that includes permutations that neglect publication date, this restricted null captures the dynamics of publication in which a new work can unexpectedly change the information space. We are unable to resolve publication dates to less than a year; this occasionally implies an ambiguity in creation date for texts in the corpus that have the same year of publication. To solve this, we average our results  over all possible within-year orders.

We compare the production and consumption of texts by considering only the texts Darwin recorded himself as reading. This allows us to make a direct comparison between the average surprise of the publication order and reading order within that set. But the production of these texts, of course, occurs in a much larger context, and it is reasonable to ask about the books that Darwin considered reading but did not, or an even more complete representation of the state of Victorian science constructed using all the scientific books available to him in Kent and London during these years. Here we focus on the books he chose to read, and the space he actually explored, partly for practical reasons and partly because we are interested in the decisions he made to order the texts, rather than the decision of whether to read them at all.

\subsection{Bayesian Epoch Estimation}
In the foraging literature, individuals are often assumed to persist in sustained periods of either exploration or exploitation. We call this an \emph{epoch}. We are particularly interested in whether or not these epochs align with important events in Darwin's life. By using Bayesian models to infer epoch breaks, we can determine whether the data support a qualitative interpretation of the quantitative model.

Bayesian epoch estimation (BEE) models an epoch as a Gaussian distribution of relative surprise, in either the text-to-text or text-to-past case, with fixed mean and variance. Each epoch is defined by a beginning point (which is also either the end of the previous epoch, or the start of the data), an average level of surprise, and the variance around that average. For each time-series, the model then contains $3n-1$ parameters, where $n$ is the number of epochs.

Epoch switches are independently selected for the text-to-text and text-to-past measures. Each transition can then be interpreted as a change in Darwin's exploration and exploitation behavior at the local or global level. When a new epoch has higher average surprise than the one before, for example, we can understand Darwin as moving to a more exploration-based strategy.

Our one externally-set parameter is the total number of epochs, $n$. As $n$ rises, it becomes easier and easier to fit the data; at some point, we encounter the over-fitting problem, and in the extreme case we have as many parameters as we have data points. For each choice of $n$, our model returns a likelihood: the probability that the observed data were generated by the (best fitting) choice of parameters for that model. As $n$ rises, so does the log-likelihood.

To determine the best-fit number of epochs, we use a simple model-complexity penalty, Akaike Information Criterion (AIC) \citep{Akaike1974}, to verify that the selected model is preferred, despite the addition of new parameters. The AIC penalizes a model's log-likelihood by the total number of parameters in the model; adding more epochs, in other words, must justify itself by a sufficiently large increase in goodness of fit. The AIC can be understood as an information-theoretic and Bayesian version of the chi-squared test, which attempts to maximize a model's predictive power \citep{burnham2003model}.

To fix attention on the longest timescales in Darwin's life, we set the minimum epoch length to five years. See \ref{S4_Text} for further discussion of the BEE model, for the likelihood space of epoch breaks for Darwin's readings, and for our AIC analysis.

\section{Results}
\subsection{Exploration and Exploitation}
Over the 647 records in our corpus, Darwin's reading order led to a below-null average surprise, where the null is the average surprise of 1,000 permutations of Darwin's reading order, constrained by each book's publication date (see Section 2.4). 

On average, the KL divergence from text to text in the corpus is 10.78 bits compared to a null expectation of 11.41 bits ($p \ll 10^{-3}$). Meanwhile, Darwin's text-to-past average surprise is 2.96 bits in the data versus 2.98 bits in the null ($p = 0.02$). Darwin's average surprise, in both text-to-text and text-to-past, is lower than expected from a null model. While our rejection of this simple null model provides little new insight into the cognitive process of a reader's decision-making---which we expect to have some correlation from book to book---it is a crucial test of the sensitivity of our methods themselves.

A surprise-minimizing path is one that orders the texts so as to minimize the total sum of text-to-text, or text-to-past, surprise. Finding this shortest path amounts to a variant of the ``traveling salesman problem'', which is famously difficult to solve (reviewed in \citet{cook2011}). The greedy shortest path algorithm attempts to approximate the surprise-minimizing path by starting with the first text that Darwin read, and choosing as the next one to read the one with smallest KL divergence from the first, and so on, minimizing at each step either the text-to-text, or text-to-past KL depending on which quantity one is interested in.

While Darwin's path is lower in surprise than the null, it is far larger than many paths that can be found: the greedy shortest-path algorithm, for example, can reduce the text-to-text average surprise to 2.11 bits and text-to-past average surprise to 2.97 bits. Table~\ref{table-steps} shows the raw local text-to-text and global text-to-past KL divergence data, along with the greedy shortest path single-visit traversals of the KL distance matrix. 

The null, actual, and greedy shortest-path results show that Darwin has a focused reading strategy despite not following a pattern of pure surprise-minimization. Interestingly, the greedy shortest path is slightly longer than the path Darwin took in the global measure. This highlights how preference for exploration at the local scale---\emph{i.e.}, not taking the closest book in topic space at each step---can lead to an unexpectedly efficient path in the global measure. See \ref{S3_Text} for an alternative rank-order analysis reinforcing this null-model testing.

\begin{table}[t]
\begin{center}
\begin{tabular}{l|c|c}
& Local & Global \\
& text-to-text & text-to-past \\
& (bits/step) & (bits/step) \\ \hline
Darwin's order & 10.78 & 2.96 \\ \hline
Null (1,000 permutations) & $11.41\pm 0.28$ & $2.98^{+0.04}_{-0.02}$ \\
($p$-value) & $\ll 10^{-3}$ & 0.02 \\ \hline
Greedy shortest path & 2.11 & 2.97\\
\end{tabular}
\end{center}
 \caption{\emph{Exploration habits}f. Average text-to-text (local) and text-to-past (global) KL Divergence (bits/step) over the reading path. Text-to-past KL is lower, as Darwin's reading spreads out to cover the topic space and lowers the information-theoretic surprise of subsequent books. Darwin's reading strategy is simultaneously more exploitative than would be expected of a random reader while also not following a strategy of pure surprise-minimization. Note that Darwin's order displays lower global surprise than the greedy shortest path, which demonstrates that selecting the next most similar book is not the best overall strategy for minimizing average global surprise over time. \label{table-steps}}
\end{table}

\subsection{Readings over Time}

\begin{figure}[p]
\includegraphics[width=\textwidth]{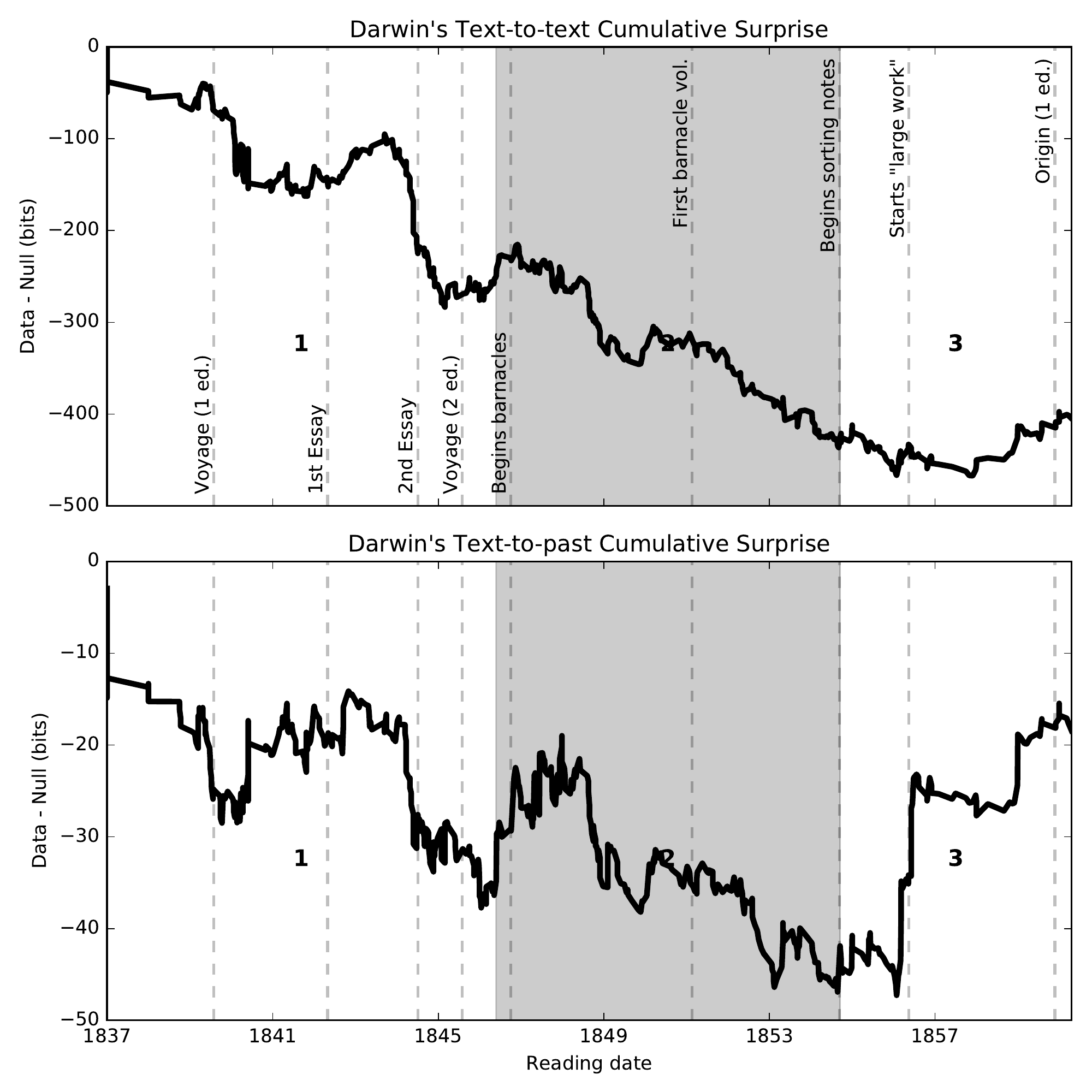}
\caption{{\bf Epochs of exploration and exploitation in Darwin's reading choices}. Text-to-text (top) and text-to-past (bottom) cumulative surprise over the reading path, in bits. More negative (downward) slope indicates lower surprise  (exploitation); more positive (upward) slope indicates greater surprise (exploration). The three epochs, identified by an unsupervised Bayesian model, are marked as alternating shaded regions with key biographical events marked as dashed lines and labeled in the top graph. The first epoch shows global and local exploitation (lower surprise). The second epoch shows local exploitation and global exploration (increased surprise, in text-to-past only). The third epoch shows local and global exploration (higher surprise in both cases).}
\label{fig:kl_over_time}
\end{figure}

While Darwin is on average more exploitative, this is not necessarily true at any particular reading date. Darwin's surprise accumulates at different rates depending on time, as can be seen in Fig.~\ref{fig:kl_over_time} for the text-to-text case (top panel) and the text-to-past case (bottom panel). These figures plot the cumulative surprise relative to the null, so that a negative (downward) slope indicates reading decisions by Darwin that produce below-null instantaneous surprise (exploitation). Conversely, a positive (upward) slope indicates decisions that are more surprising than the null (exploration). 

Over the entire corpus, as we know from the previous section, Darwin's cumulative surprise is below the null expectation, showing an overall bias towards both local and global exploitation. Tracking the slopes in these charts over time, however, allows us to see how Darwin moves between low-surprise and high-surprise choices on a range of timescales. The interaction of these decision rules at the text-to-text and text-to-past levels characterize Darwin's behavior. 

\subsection{Individual and Collective}
\begin{figure}[p]
\includegraphics[width=\textwidth]{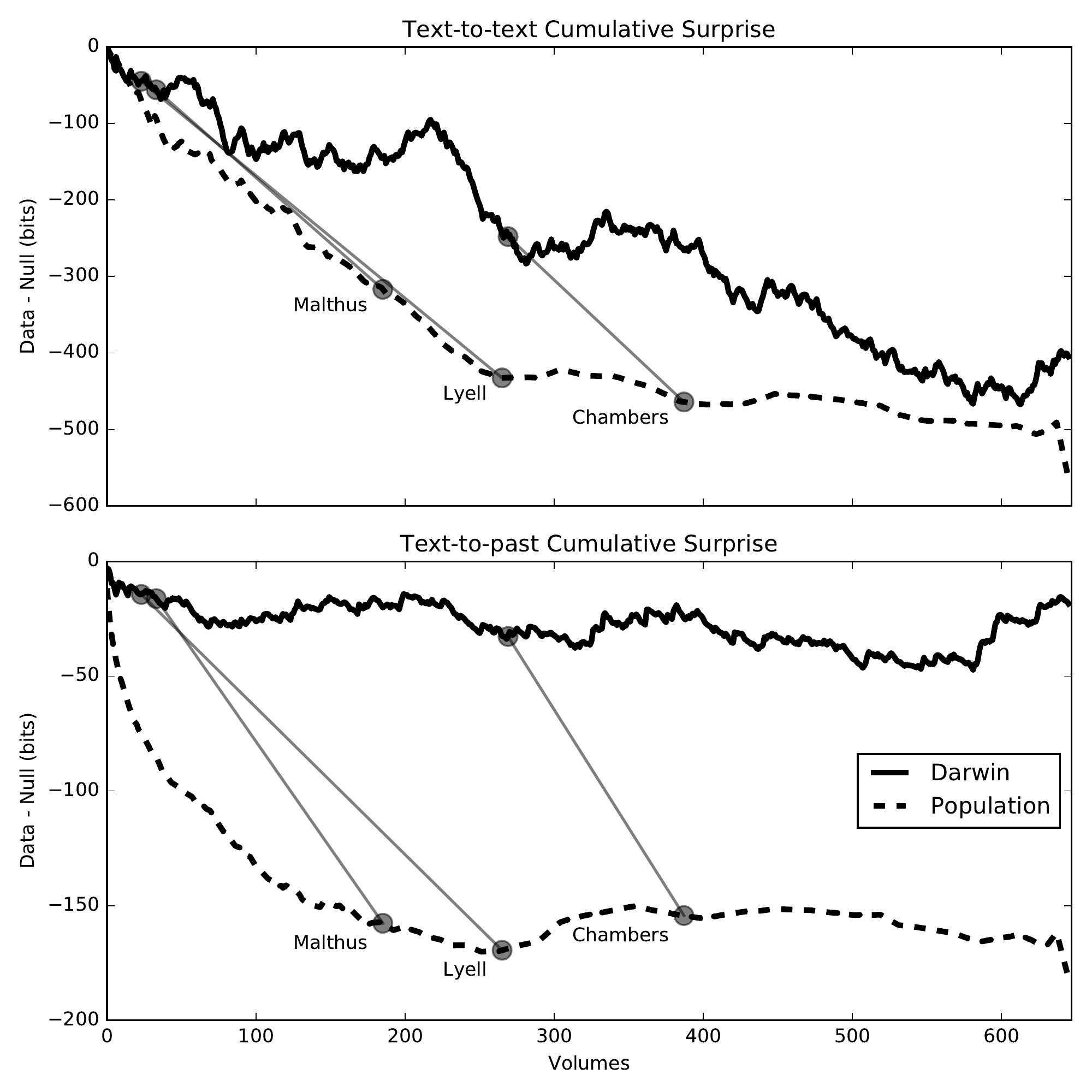}
\caption{{\bf Darwin's reading order more exploratory than the culture's production}. Text-to-text (top) and text-to-past (bottom) cumulative surprise over the reading order (solid) and the publication order (dashed). More negative (downward) slope indicates lower surprise (exploitation); more positive (upward) slope indicates greater surprise (exploration). In both cases, Darwin's cumulative surprise is higher than the publication order; in the second case, very significantly so. We mark the positions of three biographically significant books: Charles Lyell's \textit{Principles of Geology} (3rd ed., 1837; read in 1837), Thomas Malthus's \textit{An Essay on the Principle of Population} (1803; read on October 3, 1838), and Robert Chambers's \textit{Vestiges of the Natural History of Creation} (1844; read on November 20, 1844). Darwin's juxtaposition of Lyell and Malthus, for example, is characteristic of how Darwin's reading strategies reordered the products of his culture.}
\label{fig:culturalsurprise}
\end{figure}
While many studies see scientific innovations as following large-scale cultural trends \citep{Sun2013}, individuals can also be understood as ahead of their time, pursuing connections and ideas before they are recognized by the culture as a whole \citep{Johnson2010,Bliss2014}. By ordering Darwin's readings by publication date, rather than reading date, we see how the culture gradually accumulates and assimilates content. We then compare how the culture produced these texts to how Darwin, in his reading, consumed them.

Whether surprise is higher in the reading order or the publication order is a substantive empirical question. There are good reasons to imagine that the reading order will be lower in text-to-text and text-to-past surprise than the publication order---but equally plausible reasons for the opposite.

Consider, first, the idea that the reading order surprise is the lower of the two. Informally, this would suggest that reader has been at least partially successful at reordering the large number of texts into a more systematic sequence, finding similarities between chronologically distant texts and making sense of a disordered, distributed discovery process. A reader like Darwin, particularly early in his career when benefiting from more senior mentors, might be expected to attempt to do something similar.

However, equally reasonable arguments work for the opposite direction, where reading order is more surprising than the publication order. In this case, the reader is ``remixing'' the products of the past, sampling texts from different periods in such as way as to juxtapose thematically distant readings. While society accumulates themes gradually, in an exploitation regime, the reader is in an exploration regime, bringing them into unexpected contact.

While both possibilities appear \emph{a priori} plausible, the data decisively favor the second. Figure~\ref{fig:culturalsurprise} shows the text-to-text and text-to-past cumulative surprise for Darwin's reading order (solid line) compared to the publication date order (dashed line). Since volumes are published and read at different times, the $x$-axis is now ordinal (\emph{i.e.}, by position in the reading or publication sequence), rather than temporal (\emph{i.e.}, by date read or published). This allows us to compare his reading order to the publication order independent of time.

Compared to Darwin's reading practices, cultural production has lower rates of surprise. While cumulative text-to-text surprise for Darwin often shows either flat or positive (above-null text-to-text surprise) slope, the publication order path is less explorative in both text-to-text and text-to-past cases. These findings, both at high levels of statistical significance ($p\ll10^{-3}$), provide strong evidence for the remixing hypothesis.

\subsection{Strategy Shifts between Biographically Significant Epochs}

Between 1837 and 1860, Darwin's three major intellectual projects are reflected in his publication history. First, he began assembling his research journals on the geology and zoology from the voyage of the \emph{HMS Beagle}. The last of these nine volumes was published in 1846. A second epoch can be dated from 1 October 1846 when, while assembling the last of his \emph{Beagle} notes, Darwin discovered a gap in the taxonomic literature concerning the living and fossil \emph{cirripedia} (or barnacles) \citep{DAR158-Oct}. After a period of intense work, he published four volumes on the taxon from 1851 to 1854. A final epoch begins with his journal entry on 9 September 1854, marking the day he began sorting his notes for a major work on species \citep{DAR158}. The revolutionary \emph{Origin of Species} was published on 24 November 1859. These dates define three intellectual epochs: (1) from the beginning of records in 1837 to 30 September 1846; (2) from 1 October 1846 to 8 September 1854; and (3) from 9 September 1854 to the end of records in 1860.

We use the text-to-text and text-to-past models to characterize the exploration and exploitation of Darwin's reading behavior in each epoch. In instances where Darwin's average KL-divergence is above the null (more positive), Darwin is more exploratory. In instances where Darwin's average KL-divergence is below the null (more negative), Darwin is more exploitative. The degree to which he is in either mode is shown by the magnitude of the number. Table~\ref{table-darwin-epochs} shows these values.

Darwin's three biographical epochs are characterized by major shifts in both text-to-text and text-to-past surprise. Darwin begins, in epoch one, in an exploitation mode in both text-to-text and text-to-past. His turn to the barnacles in 1846 is marked by a shift from exploitation towards exploration at the text-to-past level (global shift to new area), and an intensification of his exploitation strategy at the local, text-to-text level (increased focus in this new area). In the third epoch, when Darwin ``began sorting his notes for Species Theory'' \citep{DAR158}, text-to-past remains in the exploration mode; text-to-text now shifts to exploration as well.
 
\begin{table}[th!]
\begin{center}
\begin{tabular}{c|ccc}
           & \emph{Beagle} writings & Barnacles  & Synthesis \\
Start date & 2 October 1836 & 1 October 1846 & 9 September 1854 \\ \hline 
Text-to-text & -0.68 & -0.96 & 0.32 \\
Text-to-past & -0.09 & -0.06 & 0.26 \\
\end{tabular}
\end{center}
\caption{{\bf Information-theoretic correlates of biographically significant events}. In this first table, we measure the relative surprise of Darwin's reading order by reference to dates derived from qualitative biographical work. The first major epoch of Darwin's intellectual life identified in this fashion corresponds to his post-\emph{Beagle} work, when his readings were mostly in natural history and geology. Both text-to-text and text-to-past surprise remain low---a regime of simultaneous local and global exploitation. The second epoch, when Darwin turns to a study of barnacles, shows an increase in text-to-past surprise (new topics; exploration) coupled with a decrease in text-to-text surprise (smaller jumps within these new topics; exploitation). The third epoch, when Darwin begins to collect his notes for his ``Species Theory'', is characterized by a rise in both text-to-text and text-to-past surprise. Now Darwin is neither repeatedly returning to well-covered topics (as in epoch one), nor turning his attention to a new, but narrow, range (as in epoch two), but rather ranging widely over new, previously understudied topics.}
\label{table-darwin-epochs}
\vspace{1cm}
\end{table}

\subsection{Unsupervised Detection of Strategy Shifts}

In addition to using Darwin's personally-specified epochs, we use a Bayesian model (Bayesian Epoch Estimation [BEE], see Methods) to estimate epoch breaks from text-to-text and text-to-past surprise alone. This process determines inflection points for Darwin's behavior without reference to outside biographical facts, allowing us to determine the extent to which the intellectual epochs identified by traditional, qualitative scholarship align with purely information-theoretic features of his reading.

For text-to-text surprise, we find the boundary at 16 September 1854 (log-likelihood relative to not having a boundary: $\Delta \mathcal{L} = 2.61$; 14 times as likely as not having a boundary). This is within 1 week of his journal entry on 9 September 1854 marking the start of his synthesis.  For text-to-past surprise, we find the boundary at 27 May 1846 ($\Delta \mathcal{L} = 6.17$; 479 times as likely). The difference between the automatically-selected date and his recorded start date on 1 October 1846 suggests the need for further investigation of what our results suggest was a period of relative uncertainty for Darwin about his research plans. 

The exploration-exploitation characteristics of these epochs are shown in Table~\ref{table-auto-epochs}. The close proximity of these automatically-detected breaks and the biographically significant epochs of the previous section confirm the central role of information-theoretic surprise in tracing the evolution of Darwin's search strategies. 

Both the text-to-text and text-to-past models make highly simplifying assumptions about the nature of Darwin's reading. The text-to-text case makes the most severe assumption of all: that Darwin's reading choices are conditional solely on the book just read. If Darwin's reading choices are strongly influenced by longer term memory (as seems likely), and it is these patterns which define the true epoch boundaries, it is natural that evidence for epoch boundaries in the text-to-text BEE model is weaker than the text-to-past case. In addition, our BEE makes the simplifying assumption that successive surprise values are independent draws from the distribution associated with that epoch.

\begin{table}[t]
\begin{center}
\begin{tabular}{c|ccc}
           & \emph{Beagle} writings & Barnacles  & Synthesis \\
Start date & 2 October 1836 & 27 May 1846 & 16 September 1854 \\ \hline 
Text-to-text & -0.78 & -0.76 & 0.21 \\
Text-to-past & -0.11 & -0.02 & 0.24 \\
\end{tabular}
\end{center}

\caption{{\bf Biographically significant events are detectable by unsupervised learning}.
Even in the absence of qualitative information about Darwin's life, our Bayesian model, based only on text-to-text and text-to-past surprise measurements finds---with only slight differences---the three historically-noted epochs of Table~\ref{table-darwin-epochs}: (1) from the start of our records in 1837 until text-to-past surprise changes from exploitation to exploration in Spring 1846, (2) from Spring 1846 until text-to-text surprise changes from exploitation to exploration in Autumn 1856, and (3) from Autumn 1856 to the end of our data, when both (local) text-to-text and (global) text-to-past selection behaviors are in the exploration state. The automatically-selected and biographical epochs agree on these characterizations, with small variance in the second epoch's start date.}
\label{table-auto-epochs}
\end{table}

\section{Discussion}
Models of cultural change often understand innovation as a multi-level combinatoric process, in which bundles of ideas are subject to cultural processes analogous to natural selection \citep{Jacob1977,Wagner2014}. These evolutionary analogies typically consider change at the population level, as new ideas are created, spread, and modified by the crowd. A variety of recent studies covering conceptual formation in science, technology, and the humanities have taken this population-level perspective, including work on the recombination of patents \citep{Youn2014}, novelties \citep{Tria2014}, and citations \citep{Garfield1979}. Sociological studies of scientific practice have investigated how disciplines \citep{Sun2013} or ``communities of practice'' \citep{Bettencourt2015} are formed.

The mechanisms driving cultural innovation at the population-level cannot, however, be fully understood without taking into account the cognitive processes that operate at the level of individual scientists. We have taken a step towards modeling these individual-level processes by studying the information foraging behavior of one preeminent scientist, using an information-theoretic framework applied to probabilistic topic models of his reading behavior. The information-theoretic measure we use to quantify surprise, KL divergence, connects both analytically and empirically to linguistics \citep{COGS:COGS12167,hale2001probabilistic,Levy20081126,light2002statistical,COGS:COGS1267,resnik1993selection} and visual search \citep{Demberg2008193,itti2009bayesian}. The LDA topic models that generate this information space also have cognitive correlates \citep{Griffiths2007}.

Our methods allow us to zoom in on Darwin's individual-level process to identify major epochs in his reading strategies. Over time, Darwin shifts towards increasing exploration as he prepares to write the \emph{Origin}. Interestingly, the overall order we find empirically---exploitation then exploration---is in contrast to many of predictions derived from mathematical accounts of how optimal agents navigate the exploration-exploitation dilemma~\citep{gittins1979bandit}. These generally predict that individuals begin with exploration (see, \emph{e.g.}, \cite{Berger-Tal2014}), shifting later to exploitation as they gain information about the environment. 

Our results may be consistent with the standard models, if Darwin's early exploitation was guided by an earlier exploration phase prior to 1837. Or, it may be the case that the early phase of exploitation was necessary for Darwin to gain sufficient abilities or confidence to explore in a reliable fashion later. Finally, it is worth noting that some evidence in favor of these standard models can still be found in the short-term switch towards greater exploitation at the text-to-text surprise from the first to the second epoch. This suggests that these models may be useful at shorter timescales in an individual's life.

Our use of Charles Darwin allows us to validate our methods by reference to the extensive qualitative literature on his intellectual life. Having presented a general information-theoretic framework for describing the exploitation-exploration trade-off, we can now look at other searchers to see if other strategies exist for managing the exploitation-exploration trade-offs. Expanding these results beyond the Darwin test case will be essential to providing new empirical constraints on theories of how individuals explore the cultures of their time.

Reading records exist not only for elites, like Charles Darwin, but also for the general public. The maintenance of personal reading diaries is widespread, as evidenced by the more than 30,000 records in the UK Reading Experience Database (1450--1945),\footnote{\url{http://www.open.ac.uk/Arts/reading/}} and the 50 million registered users of Goodreads.\footnote{\url{https://www.goodreads.com/about/us} (accessed 2016 July 14)} Further studies will be accelerated by advances in information retrieval techniques to find the full text for each entry in these reading diaries. These large-scale surveys can be complemented by in-depth studies of the ``commonplace books'' left by historical figures. These books record quotes, readings, and interactions that may become useful in their later intellectual life, and include Marcus Aurelius~\citep{marcus}, Francis Bacon~\citep{Bacon1883}, John Locke~\citep{Locke1706}, and Thomas Jefferson~\citep{wilson2014jefferson}.

Our method also allows us to compare the individual and the collective. We have found, in particular, that Darwin followed a path through the texts that was more exploratory than the order in which the culture produced them. Our work reveals an important distinction between these two levels of analysis; underneath gradual cultural changes are the long leaps and exploration comprising an individual's consumption, combination, and synthesis. 

Our cognitive analysis of these records builds upon decades of archival scholarship and innovations in the digital humanities. Darwin's industry extends beyond the bounds of the data we use here, however. During the \emph{Beagle} voyage, he kept a library of 180 to 275 titles \citep{DCPBeagle}. His retirement library contains 1,484 titles \citep{rutherford1908}. Darwin's handwritten marginalia in 743 of these books is currently being digitized by the Biodiversity Heritage Library. This retirement library contains many texts not included in our study of his reading records, which only last through 1860. Finally, an extensive network of correspondents also contributed to Darwin's knowledge. The Darwin Correspondence Project\footnote{\url{https://www.darwinproject.ac.uk/}} contains over 15,000 letters to and from Darwin before 1869. A complete understanding of his information foraging will necessarily seek to understand this separate social process. %

Darwin's sustained engagement with the products of his culture is remarkable. He averaged one book every ten days for twenty-three years, including works of fiction and foreign-language texts which are not part of the present analysis. For some months in our data, Darwin appears to be reading one book every two days, a fact even he was astonished by:
\begin{quote}
When I see the list of books of all kinds which I read and abstracted, including whole series of Journals and Transactions, I am surprised at my industry. \\ \small{--- \emph{Autobiography of Charles Darwin}, p. 119\nocite{DarwinAutobiography}.}
\end{quote}

\noindent Darwin not only consumed information, it consumed him. In the words of Herbert Simon: ``what information consumes is rather obvious: it consumes the attention of its recipients'' (\citeyear{Simon1971b}). Even the most ambitious individuals must confront and manage the limits of their own biology in allocating attention.

Standard theories for how individuals balance the exploration-exploitation tradeoff draw on classic work in the statistical sciences~\citep{gittins1979bandit} and machine-learning~\citep{Thrun92c}, and often focus on determining mathematically optimal strategies for different environments~\citep{Cohen2007}. Our work provides both new tools for the study of how individuals in the real world approach these problems, and new results on an exemplar individual. 

\section{Conclusion}
Charles Darwin's well-documented reading choices show evidence of both exploration and exploitation of the products of his culture. Rather than follow a pure surprise-minimization strategy, Darwin moves from exploitation to exploration, at both the local and global level, in ways that correlate with biographically-significant intellectual epochs in his career. These switches can be detected with a simple unsupervised Bayesian model. Darwin's path through the books he read is significantly more exploratory than the culture's production of them.

To what extent the patterns we identify in Darwin, and his relationship to culture as a whole, hold for other scientists in other eras is an open question. The development of an individual is in part the history of what they choose to read, and it is natural to ask what patterns these choices have in common. 

The methods we have developed and tested here represent the first application of topic modeling and cognitively-validated measures, such as KL divergence, to a single individual. These domain general methods can be used to study the information foraging patterns of any individual for whom appropriate records exist, and to look for common patterns across both time and culture.

\section*{Acknowledgments}
We thank Peter M. Todd for extensive comments on a draft of this manuscript, as well as David Kaiser, Caitlin Fausey, Vanessa Ferdinand, and Jim Lennox for their feedback on presentations of this work. We also thank the audiences at various locations where the ideas in this paper were presented, especially within the Indiana University Cognitive Science Program.  JM and SD thank the Santa Fe Institute for their hospitality while this work was completed. We thank Tom Murphy for assistance with corpus curation and Robert Rose for programming assistance. Tools for corpus preparation and modeling were produced by Robert Rose and JM while supported by the National Endowment for the Humanities Digging Into Data Challenge (NEH HJ-50092-12, CA, co-PI). JM and CA were supported by an Indiana University (IU) Office of the Vice Provost for Research (OVPR) Faculty Research Support Program (FRSP) Seed Funding Grant and Bridge Funding Grant. JM was also supported by an IU Cognitive Science Program Supplemental Research Fellowship. Finally, we thank the anonymous reviewers and the editor for their constructive comments and honest inquiries.







\newpage

\appendix
\renewcommand\thefigure{S\arabic{figure}}
\renewcommand\thetable{S\arabic{table}}
\setcounter{section}{0}

\section*{Supporting Information}
This supporting information provides additional information about the corpus curation, null model justification, alternative KL analyses, details on Bayesian Epoch Estimation, and a final section on model robustness.
The first section presents detailed characterization of Darwin's reading corpus, preparation methods, and software.
We next introduce the Kullback-Leibler Divergence measure (KL) used in our analyses, and we discuss two interpretations of the measure, from information theory and from Bayesian statistics.
We then provide further justification of the null model, noting that any representation of Victorian science will be impoverished.
Next, we present a more detailed analysis of the greedy shortest-path  through Darwin's texts, showing that he indeed does not follow a surprise-minimization strategy. We also show the rank-order distribution of each move in Darwin's reading order compared to the null models, indicating that while Darwin does not follow pure surprise minimization, he does select the nearest neighbor more often than chance.
Finally, we explicitly detail the Bayesian epoch estimation and further justify the independent selection of epoch break points. We show the AIC analysis, and then repeat the entire analysis for 3 alternative models with $k=20,40,60$ topics.

\section{Corpus Characterization}
\label{S1_Text}

\begin{figure}[b!]
\begin{center}
\includegraphics[width=0.46\textwidth]{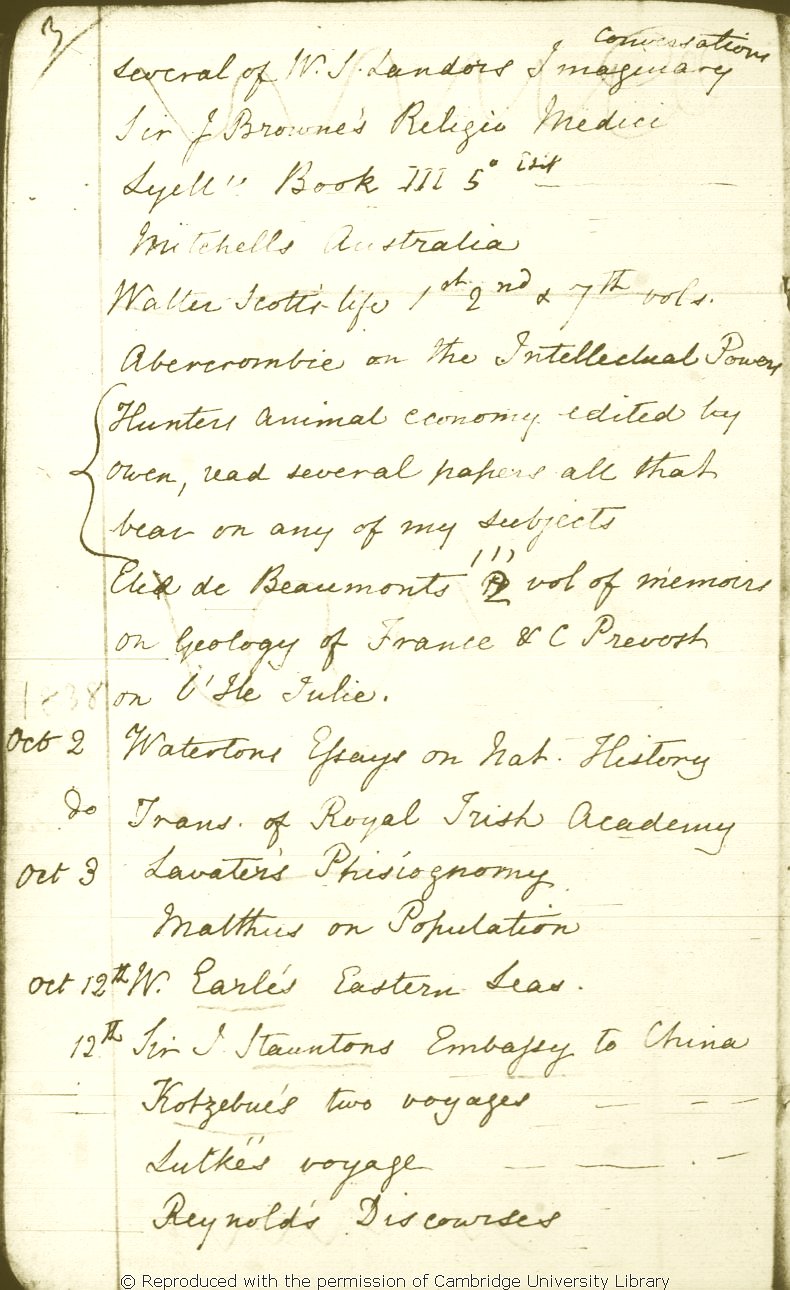}
\end{center}
\caption{\emph{Darwin's Reading Notebooks}. Page 3a of Darwin's first notebook (DAR 119), during which he began to track the exact dates. Note the reading of Malthus's \emph{On Population} on October 3, 1838. Photo courtesy of Cambridge University Libraries.}
\label{fig:notebook-page}
\end{figure}

Despite our use of a digital library, it is important to remember that books are originally physical artifacts (see Fig.~\ref{fig:notebook-page}). Victorian publishing practices often spread a single title over multiple volumes for portability and ease of use. In this paper, we use \emph{volume} to refer to each physical artifact. Each individual entry of Darwin's notebooks is referred to as a \emph{title}. In the case of books, a \emph{title} gathers together one or more volumes. In the case of journal articles, a \emph{title} is merely a subpart of a particular volume. We model at the level of a \emph{catalog record}, which corresponds to a \emph{title} in almost all cases, except for journals, where it corresponds to the aggregate of all issues listed as read across entries in the notebooks. 

Darwin also read French, German, and Latin texts. We reduced the corpus to English-only to reduce cross-linguistic effects in the model \citep{BoydGraber2009}. Additionally, we focused only on non-fiction texts. An examination of the influence of fiction on Darwin is a topic for further exploration.

There are 647 catalog records and 1057 volumes corresponding to the 665 titles modeled in this study. Some volumes in the corpus alignment were unable to be matched to the exact edition listed by the DCP, and thus there is occasionally a difference between the volume Darwin read and the volume whose text we use for topic modeling. Table~\ref{table-corpus-stats} shows the summary of the items which were located and remain missing.

\begin{table}[h!]
\begin{center}
\begin{tabular}{l|rr|r}
& Located & Non-located & Total \\ \hline
\textbf{Total} & \textbf{811} & \textbf{104} & \textbf{915} \\
- Fiction & - 79 & - 1 & - 80  \\
- Non-English & - 63 & - 85 & - 148 \\ \hline
\textbf{English Non-fiction} & \underline{\textbf{665}} & \textbf{22} & \textbf{687} \\ \hline
\end{tabular}
\end{center}
\caption{\emph{Corpus Composition:} Composition of the Reading List in terms of fiction, non-fiction, English, and non-English texts. Located titles refers to the number identified in the HathiTrust (\url{http://hathitrust.org/}), Internet Archive (\url{http://archive.org/}), and Project Gutenberg  (\url{http://gutenberg.org/}). Non-located texts were unavailable in the HathiTrust, Internet Archive, or Project Gutenberg as of December 1, 2015.}
\label{table-corpus-stats}
\end{table}

Our publication dates are those listed by the Darwin Correspondence Project (DCP); the DCP uses the publication date of the volume, if found in Darwin's library, otherwise the date of first publication. The reading order is determined by dates listed in the reading notebook. When multiple titles are listed at a particular date, we use their natural ordering in the notebooks --- titles written at the top of the page are assumed to be read before those at the bottom.

We use the InPhO Topic Explorer \citep{Murdock2015} for tokenization and modeling of texts. First, plain-text editions downloaded from the HathiTrust, the Internet Archive, and Project Gutenberg are normalized by merging cross-line hyphens into single words, normalizing into ASCII using the Python library Unidecode, removing all words containing punctuation and numerals (often due to OCR errors), and lower-casing all words. Then, words appearing in the English stopwords corpus from the Natural Language Toolkit (NLTK) \citep{nltk} are removed. Finally, words occurring less than 30 and more than 15,000 times were excluded from the corpus. After this pre-processing, the corpus consisted of 40,822,136 tokens drawn from 77,611 unique tokens. We made no attempt to apply stemming or clean up OCR errors, other than the filtering of words occurring fewer than 30 times.

\begin{figure}
\begin{center}
\includegraphics[width=0.95\textwidth]{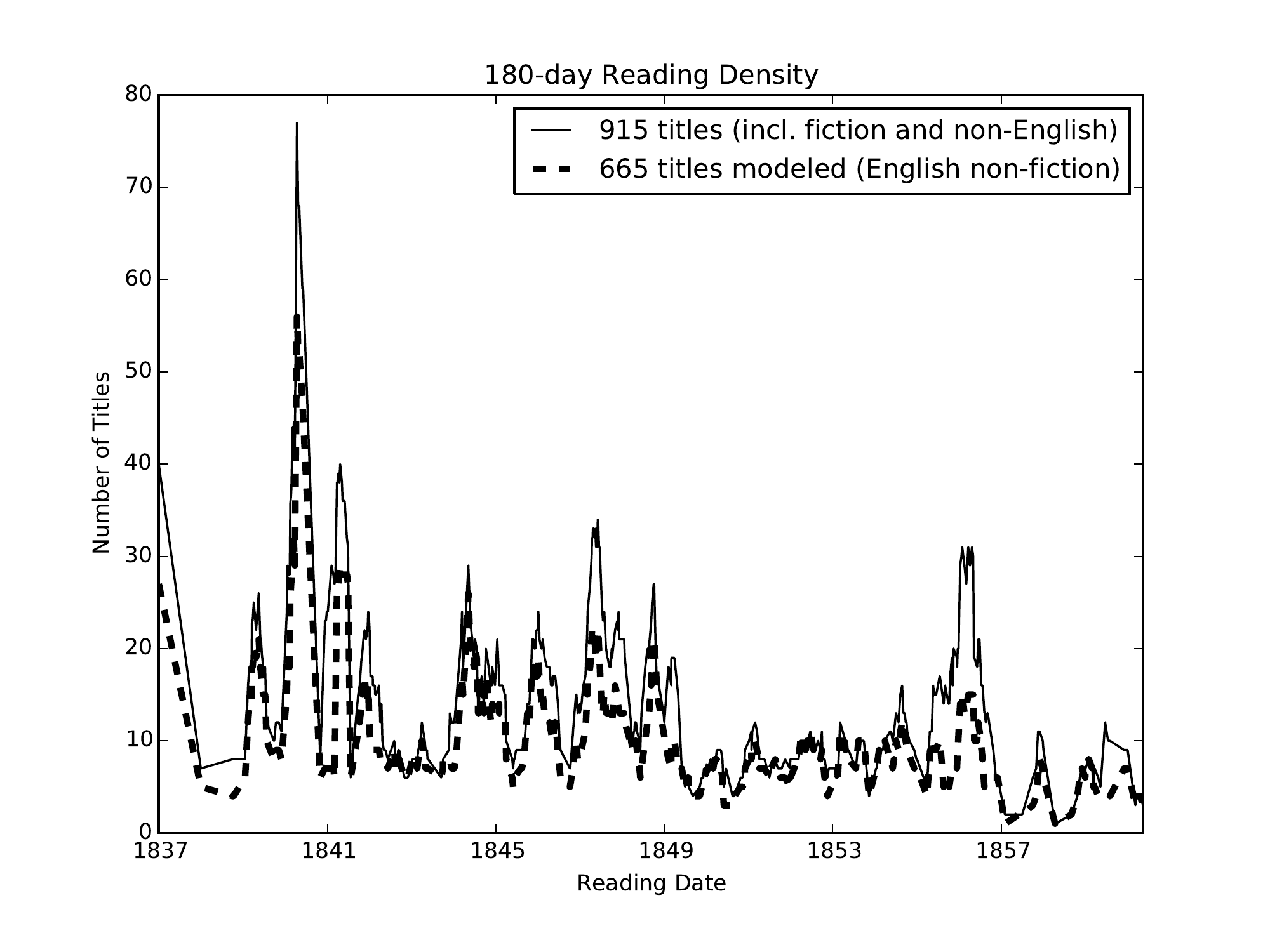}
\end{center}
\caption{\emph{Reading Density} -- Reading density, smoothed over a 6-month window.  The dashed line shows the 665 titles here modeled, while the thin solid line represents all 915 titles in the reading notebooks.}
\label{fig:reading-density}
\end{figure}

Fig.~\ref{fig:reading-density} shows the density of Darwin's readings modeled. Notice the large jump in 1840 corresponds to a period when he was reading entire series of journals, each article of which was a separate title in his notebook. Also, note that Fig.~\ref{fig:reading-density} shows both the density of the selection modeled and the entire reading notebook list.

\begin{figure}
\begin{center}
\includegraphics[width=\textwidth]{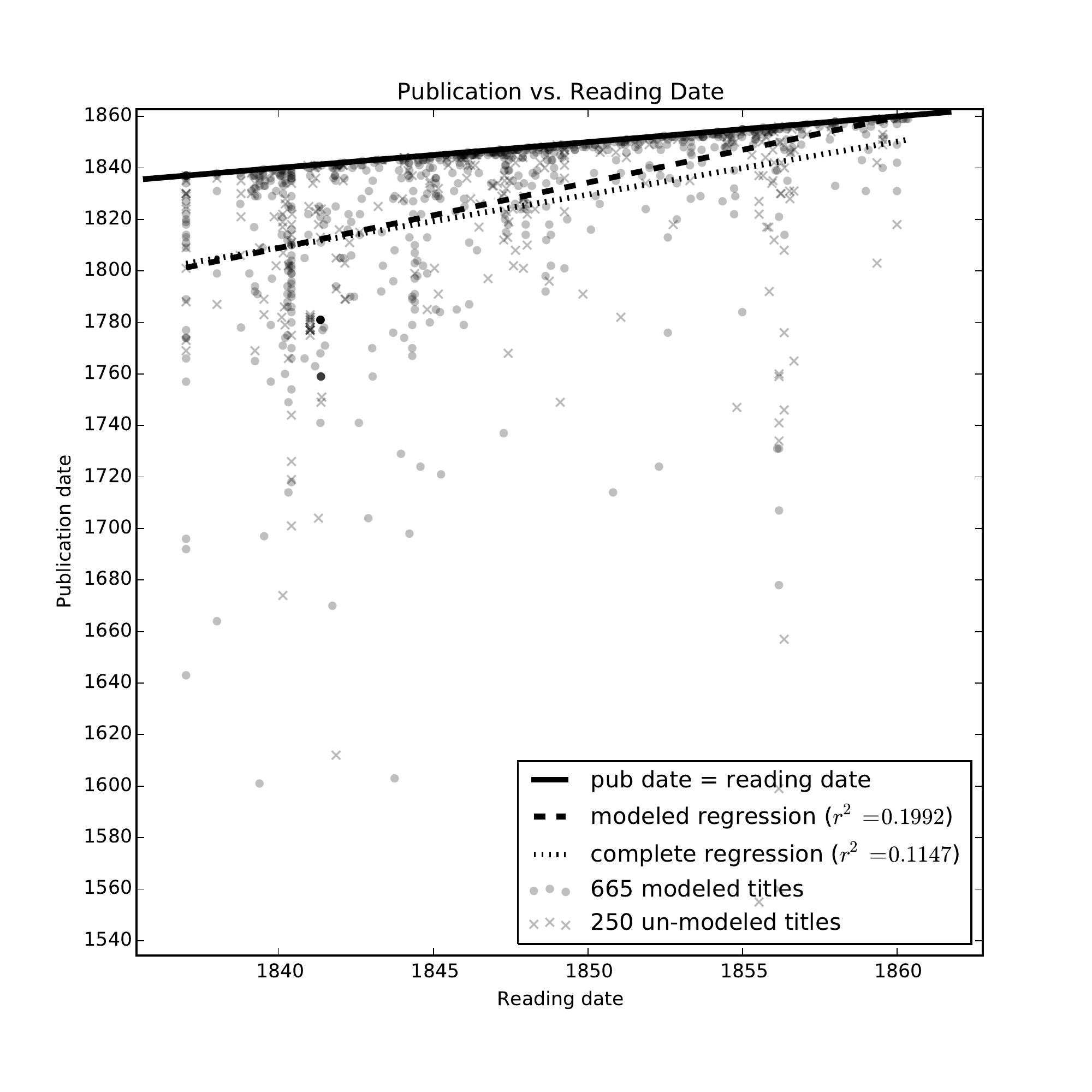}
\end{center}

\caption{\emph{Publication vs. Reading Dates} -- Scatter plot of the publication and reading dates of the titles in Darwin's reading list. The 665 modeled titles are shown with dots, while the remaining 250 titles are shown as \texttt{x}s.  The solid line indicates when the reading date and publication date are equal. The dashed line indicates a linear regression over the dots ($r^2 = 0.1992$), and the dotted line indicates a linear regression over the dots and \texttt{x}s combined ($r^2 = 0.1147$). The appearance of older materials in 1856-57 corresponds to Darwin's literature review of pigeon breeding, conducted as a case study in artificial selection and included in \emph{The Origin of Species}.}
\label{fig:read-pub-dates}
\end{figure}

Fig.~\ref{fig:read-pub-dates} indicates that as Darwin's readings progress he begins reading more recently published, contemporary sources. We also show the regression for the un-modeled texts, showing that his total reading also progressed toward contemporary sources, although at a slower rate.

\section{An Introduction to Kullback-Leibler Divergence}
\label{SKL_Text}

Kullback-Leibler Divergence (KL) will be our a measure of ``cognitive surprise'' in higher-order linguistic search processes: as we say in the main text, a way to measure the extent to which a learner encounters the unexpected in ways that require them to alter the way they represent the world. To see why KL is particularly useful for this, it is worth returning to one of the canonical interpretations of information theory: as a way to quantify the efficiency with which a process can be represented. Consider, for example, the children's game ``Twenty Questions'': one player (the ``parent'') chooses a noun at random. The other player (the ``child'') must then attempt to guess the noun, by asking a series of questions that the parent can only answer ``yes'' or ``no''.

One way to play the game is, of course, for the child herself to guess nouns at random: ``is it a car?'' ``is it a tree?''. But there are usually better ways to play the game: the child might begin by asking ``is it something that can be dead or alive?'', narrowing down the possibilities more quickly.

It is also clear that which questions are better to ask depends upon how the parent chooses the noun to begin with: the probability $p_i$ for each noun $i$. If, for example, a (somewhat lazy) parent chooses the word ``car'' 90\% of the time, ``is it a car?'' will be a particularly effective opening question for the child. Furthermore, if the child is playing well, she will use a branching script: which question to ask next will generally depend upon the answers to all the questions that came before; if we know that the noun is something that can be dead or alive, we know to skip the car questions.

How efficient can the child be? More formally, what is the average number of yes/no questions the child has to ask, if she is using one of the best possible scripts? Remarkably, this is related to the Shannon entropy of the probability distribution $\vec{p}$:
\begin{equation}
H(\vec{p})=-\sum_{i=1}^N p_i \log_2{p_i}.
\end{equation}
In particular, the average number of questions $Q$ that the child has to ask, if she is using an optimal script, lies somewhere between
\begin{equation}
H(\vec{p}) \leq Q \leq H(\vec{p})+1,
\label{ent}
\end{equation}
a result known as Shannon's source coding theorem, and proven in \citet{shannon}. For an informal introduction to this result, see \citet{dedeo}; the upper limit in Eq.~\ref{ent} can be made arbitrarily close to $H(\vec{p})$ if the child is allowed to play multiple games simultaneously.

Let us now imagine that the child has developed an optimal script for one of her parents. What happens when she uses this script playing against the other?

If the two parents have similar patterns of noun choice, she may do well. But if the patterns differ---if the first parent's distribution, $\vec{p}$ is very different from the second parent's distribution, $\vec{q}$---then her script could fail very badly. This is true even if there exists an optimal script for $\vec{q}$ that is just as efficient (for $\vec{q}$) as the original script was for $\vec{p}$.\footnote{If, for example, the first parent chooses from one of three words, $\{A, B, C\}$, with probabilities $\{1/2, 1/4, 1/4\}$, the child can ask ``is it A?'' as her first question, and ``is it B?'' as her second; on average, she will finish the game in 1.5 questions. Confronted with the second parent, whose probabilities are $\{1/4, 1/2, 1/4\}$, however, she will take on average $1.75$ questions before finishing the game---even though an optimal script exists that requires only $1.5$. The KL divergence approximates this coding failure (and becomes exact in certain natural limits); here, in fact, the KL divergence from the first parent to the second is precisely the coding failure, $0.25$ bits, or ``excess questions''.}

It so turns out that this coding failure can also be quantified. In particular, if the child builds an optimal script for distribution $\vec{p}$, but then encounters the parent with distribution $\vec{q}$, she will, on average, ask more questions than she needs to. The average number of excess questions is related to the KL divergence, which is is defined as
\begin{equation}
D_\mathrm{KL}(\vec{q} | \vec{p}) = \sum_{i=1}^k q_i \log_2 \frac{q_i}{p_i}.
\label{klapp}
\end{equation}
This becomes an exact measure of coding failure when, as before, the child is playing multiple, simultaneous games.

It is worth noting that KL is a directed measure: the divergence from $\vec{p}$ to $\vec{q}$ is, generically, different from the divergence from $\vec{q}$ to $\vec{p}$. If one builds a script for a $\vec{p}$ biased towards some subset of topics, but encounters a more uniform $\vec{q}$, the coding will fail more badly than if $\vec{p}$ were uniform and $\vec{q}$ highly biased. KL, in other words, is not a measure of distance between texts, but a way to quantify the ordered processing that happens as learning unfolds over time.

A second interpretation of KL divergence comes from Bayesian statistics. Imagine an agent observing a process and trying to decide whether her observations are drawn from $\vec{p}$ or $\vec{q}$, when her priors are equally split between the two possibilities. Given a single observation, of type $i$, the relative log-likelihood of distribution $\vec{q}$ compared to distribution $\vec{p}$ is just
\begin{equation}
\Delta\mathcal{L}=\log{\frac{q_i}{p_i}}.
\label{stepone}
\end{equation}
If the true distribution is $\vec{q}$, the average rate at which this relative log-likelihood increases is simply
\begin{equation}
\sum_{i=1}^N q_i \Delta\mathcal{L}.
\label{steptwo}
\end{equation}
On substituting Eq.~\ref{stepone} into Eq.~\ref{steptwo}, we recover the KL divergence equation Eq.~\ref{klapp}. This provides a second interpretation for KL: the rate at which log-evidence for the true distribution accumulates over time. When the new distribution is very different from the one that came before (``more surprising''), evidence for that difference will accumulate quickly.

\section{Null Model Justification}
\label{S2_Text}
A more complete representation of the state of Victorian science (i.e. Darwin's entire search space) would require the null model to be constructed against all the books available to him in Kent and London during these years, rather than books Darwin actually read. To construct and model such a corpus would be a monumental task, and would be circumscribed by the subsequent curatorial decisions that have shaped present access to digitized Victorian era texts. 

Fortunately, the null model based on Darwin's own reading list provides a more rigorous test of our results.  This is because text-to-text surprise over the entire set of published books is expected to be greater, thus accentuating the difference between data and null for Darwin's lower-surprise trajectory. Similarly, text-to-past surprise should be greater in a null model constructed against a broader set of books. This is because whether the prior state of the ``null'' reader based on one text or many, the model of the larger corpus provides more opportunities for long range jumps.

\section{Text-to-text and text-to-past KL}
\label{S3_Text}
Table~\ref{table-steps} shows the raw local text-to-text and global text-to-past KL divergence data, along with the greedy shortest path single-visit traversals of the KL distance matrix. While Darwin's average text-to-text surprise is lower than expected from a null model, it is far larger than many paths that can be found: a greedy shortest-path algorithm, for example, can reduce the text-to-text average surprise to 2.14 bits and text-to-past average surprise to 2.86 bits.

\begin{table}[h]
\begin{center}
\begin{tabular}{l|c|c}
& Local & Global \\
& (bits/step) & (bits/step) \\ \hline
Measured & 10.78 & 2.96 \\ \hline
Null & $11.41\pm 0.28$ & $2.98^{+0.04}_{-0.02}$ \\
($p$-value) & $\ll 10^{-3}$ & 0.02 \\ \hline
Greedy Shortest Path & 2.11 & 2.97\\
\end{tabular}
\end{center}
\caption{\emph{Exploration habits}. Average text-to-text (local) and text-to-past (global) KL Divergence (bits/step) over the reading path. Text-to-past KL is much lower, as Darwin's reading spreads out to cover topic space and lowers the information-theoretic surprise of subsequent books. Darwin's reading strategy is simultaneously more exploitative than would be expected of a random reader while also not following a strategy of pure surprise-minimization. \label{table-steps}}
\end{table}

\subsection{Rank Distribution}
\begin{figure}[t]
\includegraphics[width=\textwidth]{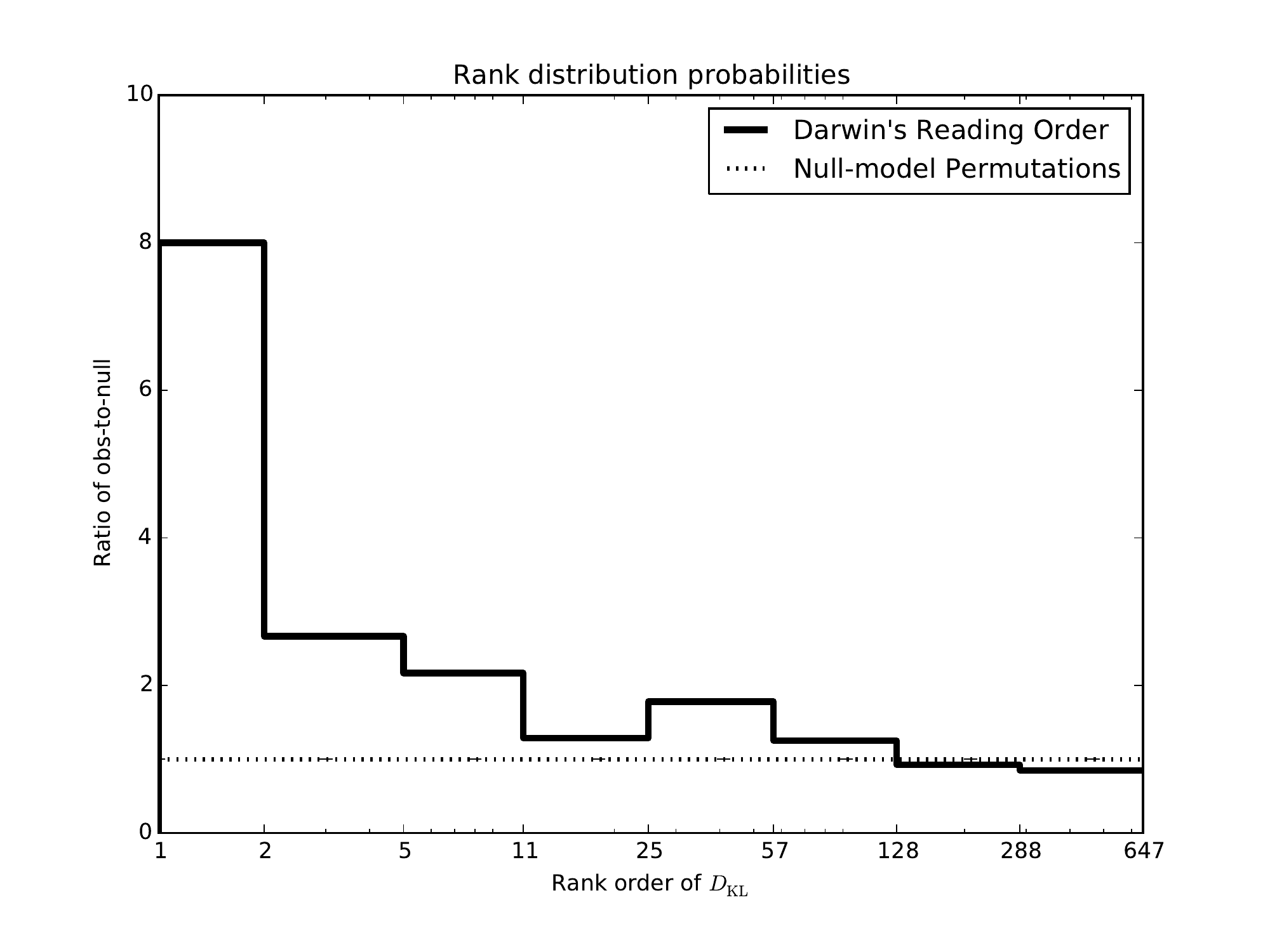}
\caption{\emph{Rank Distribution}. Rank distribution of $D_{\mathrm{KL}}(\theta_i, \theta_{i+1})$ for Darwin's reading notebooks relative to a null-model permutation of his reading order, as indicated by the dashed line, with 95\% confidence intervals shown. The lines are logarithmically binned, showing clearly that Darwin is 8 times more likely to select the nearest KL neighbor, as opposed to volumes further away, which are selected 0.85 times as likely than the null.}
\label{fig:rankdistribution}
\end{figure}

In addition to the information-theoretic measures described in the paper, descriptive statistics also capture Darwin's explore-exploit behavior. For each volume, we look at the rank of the KL divergence to the next volume by reading order compared to all other volumes in the corpus, as shown in Fig.~\ref{fig:rankdistribution}. We can compare this to a null model, as described in the Methods.

Interestingly, Darwin is 8 times more likely than the null model to pick the nearest neighbor, indicating that explorations are overall rarer than exploitations, and emphasizing that exploitations do indeed occur on a text-to-text basis.

\section{Bayesian Epoch Estimation}
\label{S4_Text}

Our generative model for Bayesian epoch estimation has $3n - 1$ parameters. There are $(n-1)$ parameters to describe the end points of the first $n-1$ epochs, and $2n$ parameters to describe the mean and variance of the text-to-text (or text-to-past) surprise within each epoch. We estimate these parameters using an approximate maximum-likelihood procedure. Within each epoch $i$, we assume the surprise is constant and Gaussian distributed with a particular mean $\mu_i$ and variance $\sigma^2_i$. We write the $3n-1$ parameters as a vector $\vec{v}$; then the distribution over $\vec{v}$ given the data $s$, equal to a list of surprises, $\{s_i\}$, is
\begin{equation}
\log P(\vec{v}|s) = \log P(s|\vec{v}) + C = -\sum^n_{i=1} \frac{(e_{i+1}-e_i-1)}{2}\left(1+\ln(2\pi \hat{\sigma^2}_i) \right) + C,
\end{equation}
where $e_i$ is the start point of epoch $i$ and $C$ depends on the prior. The start point of the first epoch, $e_1$, is fixed to be volume zero; given our conventions, $e_{n+1}$ is fixed to be the final volume plus one. The sigma estimator, $\hat{\sigma^2}_i$, is the standard maximum likelihood estimator of the variance,
\begin{equation}
\hat{\sigma^2}_i = \frac{1}{e_{i+1}-e_i-1}\sum^{e_{i+1}-1}_{k=e_i} \left(s_k-\hat{\mu}_i\right)^2,
\end{equation}
and $\hat{\mu}_i$ is defined as
\begin{equation}
\hat{\mu}_i =  \frac{1}{e_{i+1}-e_i-1}\sum^{e_{i+1}-1}_{k=e_i} s_k.
\end{equation}
To do Fisher maximum-likelihood estimation, we ignore the effect of the prior $P(\vec{v})$ on the maximum; equivalently, we do maximum a posteriori estimation and assume that $P(\vec{v})$ is flat over the region of interest.

\begin{figure}
\begin{center}
\includegraphics[width=\textwidth]{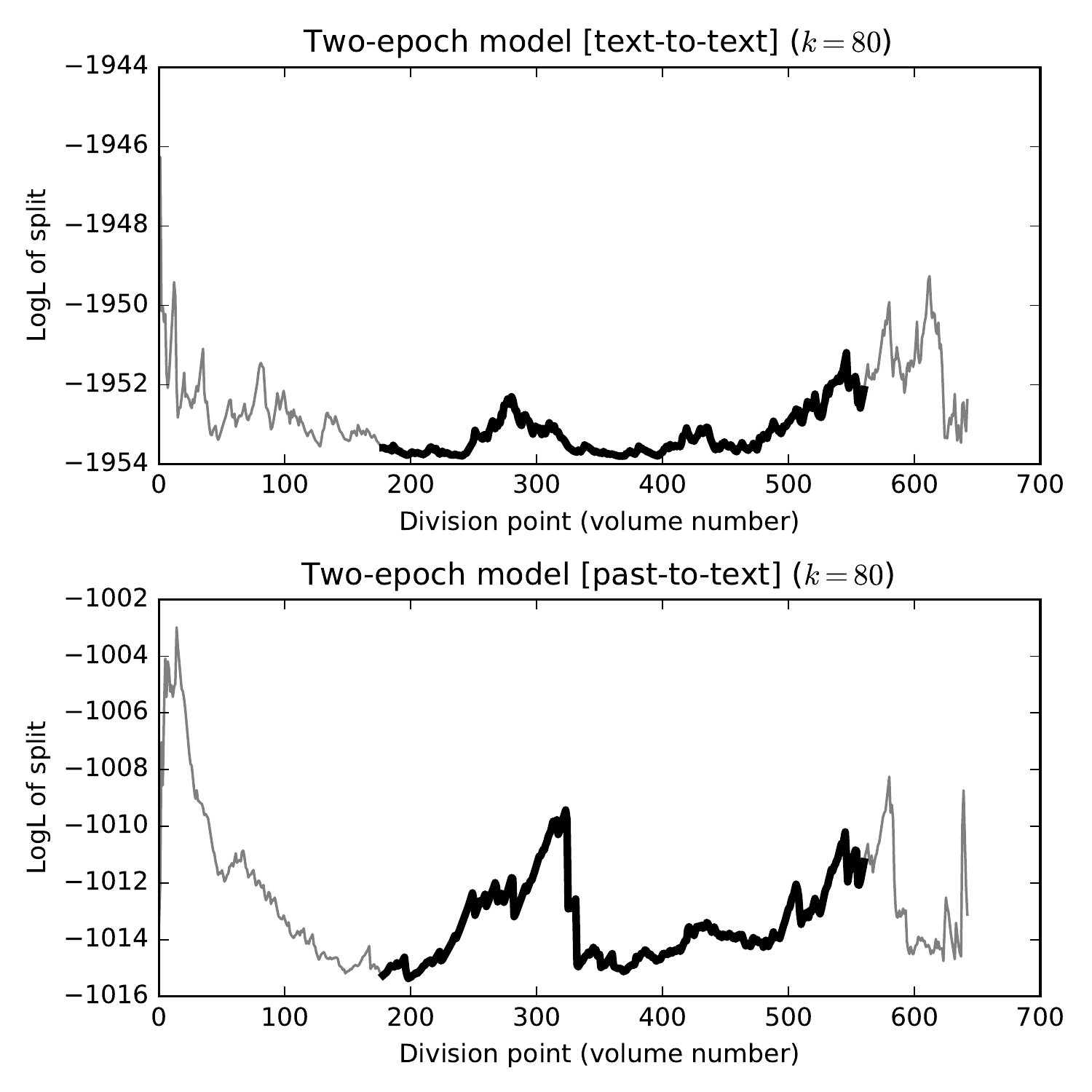}
\end{center}
\caption{\emph{Two-epoch model} -- Fisher maximum-likelihood estimation for a 2-epoch BEE model over the text-to-text and text-to-past $k=80$ models of 647 of Darwin's readings. The darker line indicates the window of the 5-year minimum epoch length. Note the phase transition at the 325th volume in the text-to-past case (bottom) and the 548th volume in the text-to-text case (top). Note also that the text-to-past case comes close to transition at the 548th volume as well, indicating the strength of the transition to exploration in the third epoch on both local and global scales.}
\label{fig:epoch-estimation}
\end{figure}

The results for a 2-epoch independent selection model are shown in Figure~\ref{fig:epoch-estimation}. Note the alternative maxima in the text-to-past model.

\subsection{Epoch Model Selection}
\label{S5_Text}
To verify that our model is not over-fitting, we use the Akaike Information Criterion \citep{Akaike1974}; we increase the number of epochs until the increase in the log-likelihood is less than the complexity penalty, equal to the number of parameters.

For the two- and three-epoch models we compare this likelihood to the a single-epoch null model of 2 parameters - mean and variance for text-to-text or text-to-past surprise over the whole data-set. Our 2-epoch model has 5 parameters. A 3-epoch model has 8 parameters.

This AIC analysis further emphasizes the relative strength of evidence for text-to-text and text-to-past epoch boundaries (see Results). Evidence for boundaries in the text-to-text BEE model is naturally weaker than the text-to-past case. The text-to-text model assumes decisions are made solely by the last read text, as opposed to longer term memory, which seems more plausible. In addition, our text-to-past BEE makes the simplifying assumption that successive surprise values are independent draws from the distribution associated with that epoch. Because memory accumulates over time, this assumption, at best, only approximates the text-to-past case.

The results of our AIC analysis are shown in Table~\ref{table:aic}.

\begin{table}[h]
\begin{center}
\begin{tabular}{r|c|c|c|c}
& Breaks & k & AIC & relative $\mathcal{L}_{AIC}$ \\ \hline
Null T2T & $[0,646]$ & 2 & 3911.61 & $1.0$ \\
1-epoch T2T & $[0,548,646]$ & 5 & 3912.38 & $0.68$ \\ 
2-epoch T2T & $[0,383,548,646]$ & 8 & 3914.67 & $0.21$ \\ \hline
Null P2T & $[0,646]$ & 2 & 2035.18 & $1.0$ \\
1-epoch P2T & $[0,325,646]$ & 5 & 2028.83 & $23.82$ \\ 
2-epoch P2T & $[0,422,547,646]$ & 8 & 2021.93 & $750.70$ \\ \hline
\end{tabular}
\end{center}
\caption{\emph{AIC Model Selection}. The likelihood for each 1-epoch selection is shown in Figure~\ref{fig:epoch-estimation}. The AIC of the independent selection for a 2-epoch model is also shown.  Note that the AIC for text-to-past selects for epoch breaks, but not for text-to-text (see Results).}
\label{table:aic}
\end{table}

\section{Model Robustness}
The ``model checking problem'' is an enduring problem for applied topic modeling \citep{Blei2012}, but recent work on selection of a ``reference model'' in the social sciences provided guidance to selecting a value of $k=80$ for the number of topics to use in our analysis \citep{Roberts2015}. More specifically, setting $k=80$ produced a set of topics subjectively deemed more interpretable than the lower values of $k$ suggested by more ``objective'' measures of model fit to data.

\begin{figure}
\begin{center}
\includegraphics[width=\textwidth]{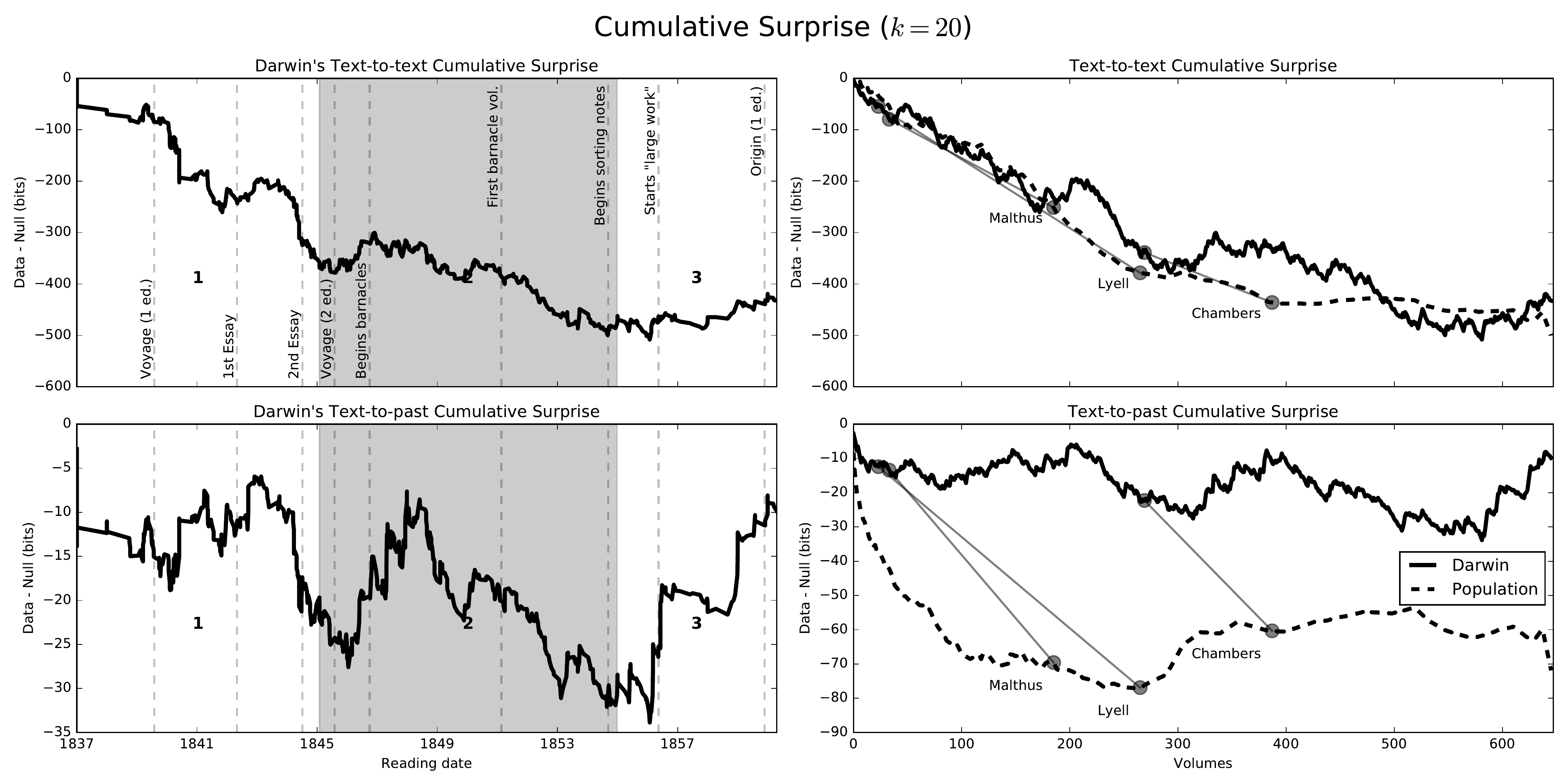}
\end{center}
\caption{\emph{Cumulative and Cultural Surprise ($k=20$)} -- Analysis of Figures 1 (left) and 2 (right) repeated for $k=20$. \emph{Left:} Average text-to-text (top left) and text-to-text (bottom left) cumulative surprise over the reading path and over the publication history, measured as the cumulative KL divergence (bits). As Darwin drops below zero in these plots, his choices are producing surprises lower than expected in the null. \emph{Right:} Average text-to-text (top) and text-to-past (bottom) cumulative surprise over the reading order (solid) and over the publication order (dashed), measured in bits.}
\label{fig:k20}
\end{figure}

\begin{figure}
\begin{center}
\includegraphics[width=\textwidth]{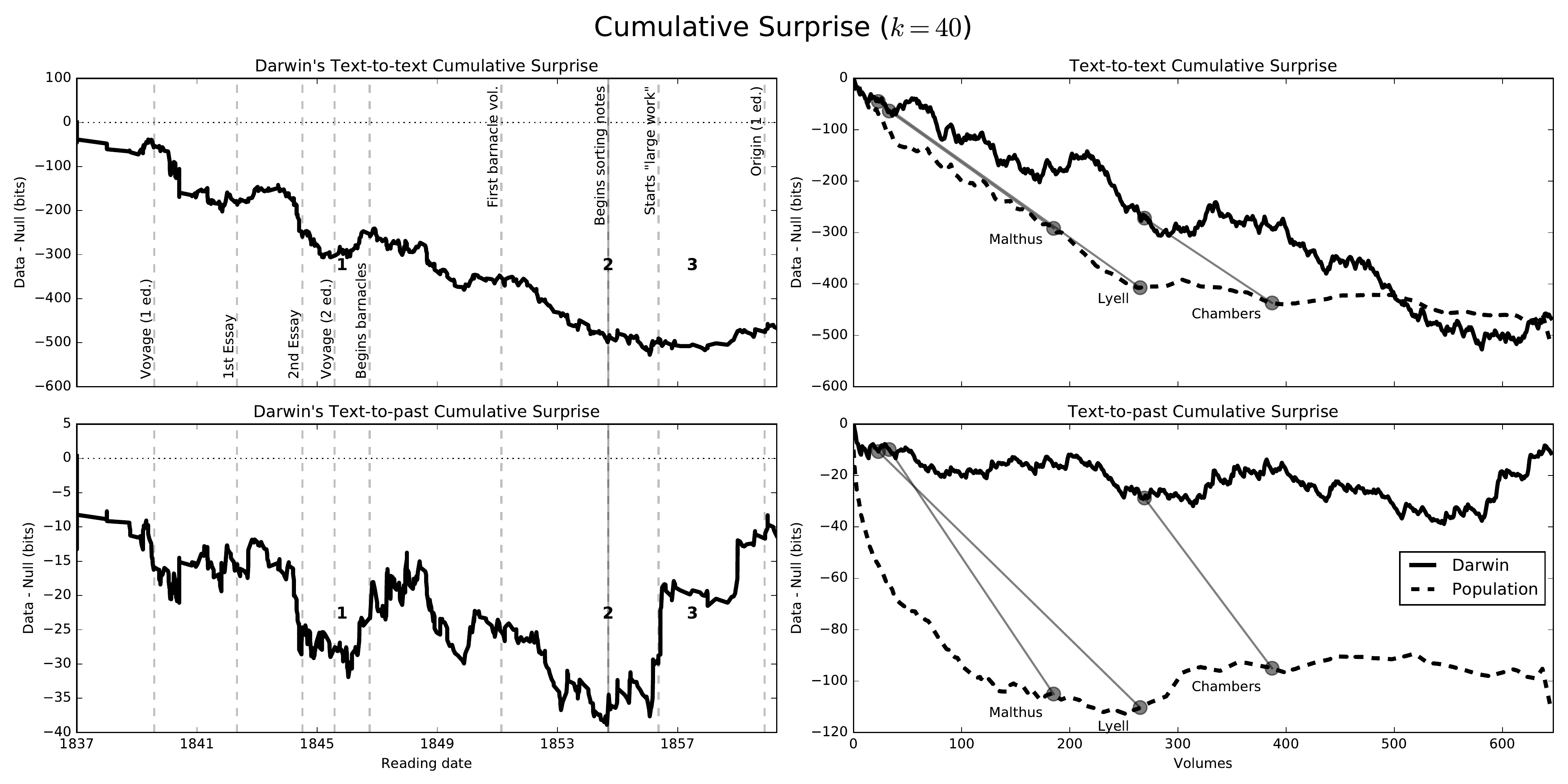}
\end{center}
\caption{\emph{Cumulative and Cultural Surprise ($k=40$)} -- Analysis of Figures 1 (left) and 2 (right) repeated for $k=40$. \emph{Left:} Average text-to-text (top left) and text-to-past (bottom left) cumulative surprise over the reading path and over the publication history, measured as the cumulative KL divergence (bits). The $k=40$ model does not select an epoch boundary in 1846, resulting in a very short 1-volume epoch. \emph{Right:} Average text-to-text (top) and text-to-past (bottom) cumulative surprise over the reading order (solid) and over the publication order (dashed), measured in bits.}
\label{fig:k40}
\end{figure}

\begin{figure}
\begin{center}
\includegraphics[width=\textwidth]{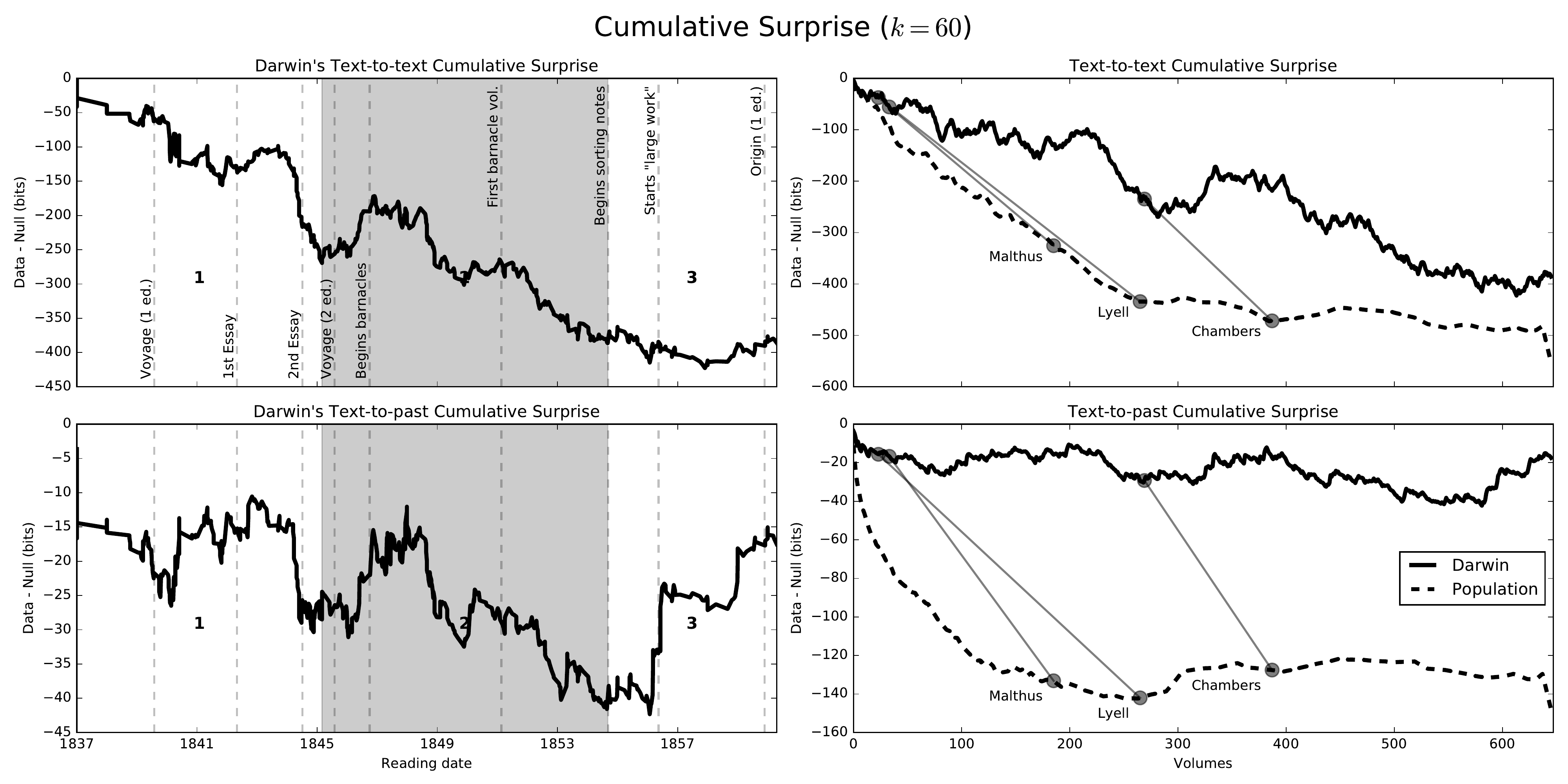}
\end{center}
\caption{\emph{Cumulative and Cultural Surprise ($k=60$)} -- Analysis of Figures 1 (left) and 2 (right) repeated for $k=60$. \emph{Left:} Average text-to-text (top left) and text-to-past (bottom left) cumulative surprise over the reading path and over the publication history, measured as the cumulative KL divergence (bits). \emph{Right:} Average text-to-text (top) and text-to-past (bottom) cumulative surprise over the reading order (solid) and over the publication order (dashed), measured in bits.}
\label{fig:k60}
\end{figure}

In addition to the $k=80$ topics results shown in the main paper, the same analyses are also shown below for $k={20,40,60}$ in Figs.~\ref{fig:k20}, \ref{fig:k40}, \ref{fig:k60}, respectively. 

Epoch breaks are also noted in the figure captions. All topic models agree on a break in 1854, but the $k=40$ model does not detect an earlier break in the text-to-past case, rather selecting volume 548. The $k=60$ model seems to be an outlier from the other values of $k$.


\begin{figure}
\begin{center}
\includegraphics[width=.31\textwidth]{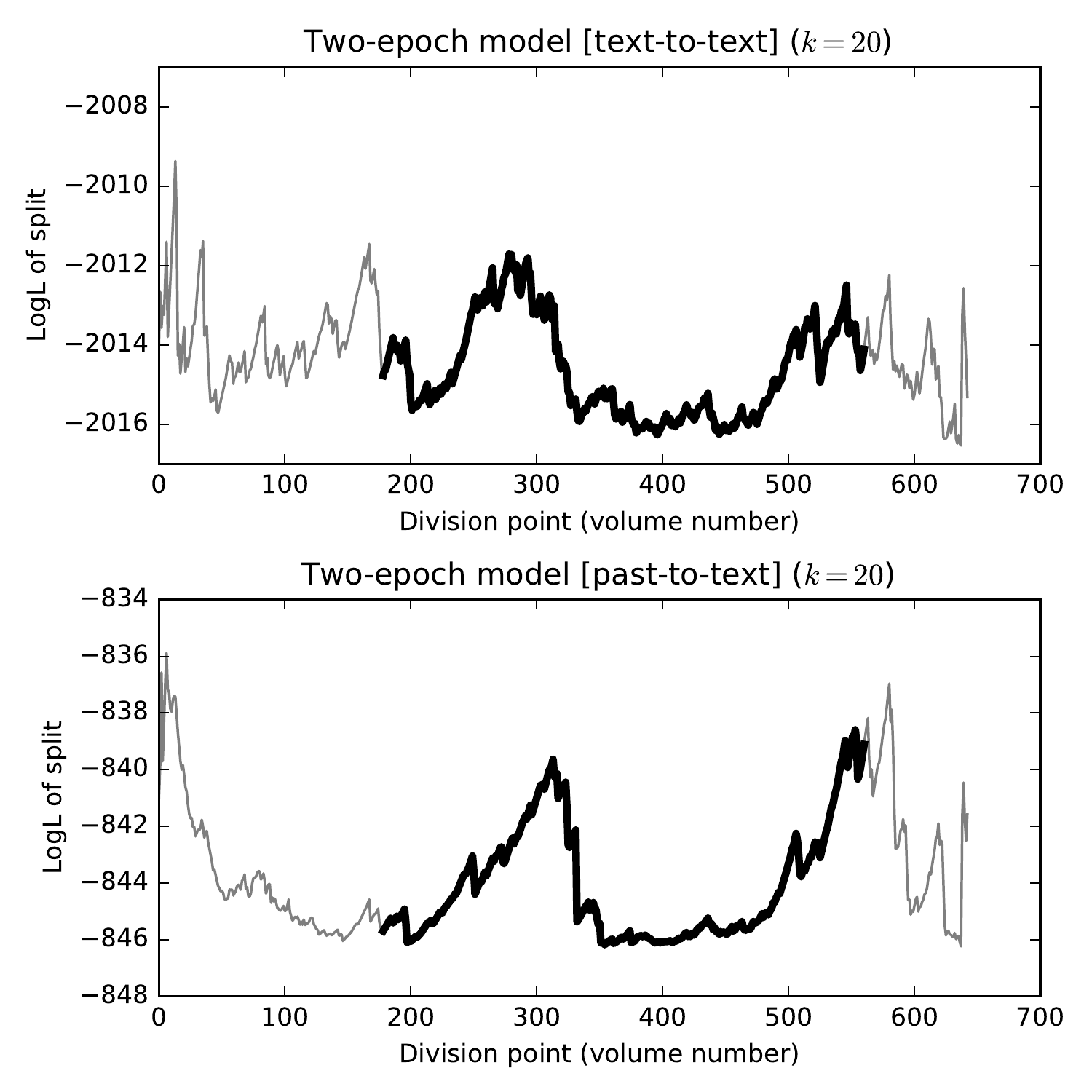}
\includegraphics[width=.31\textwidth]{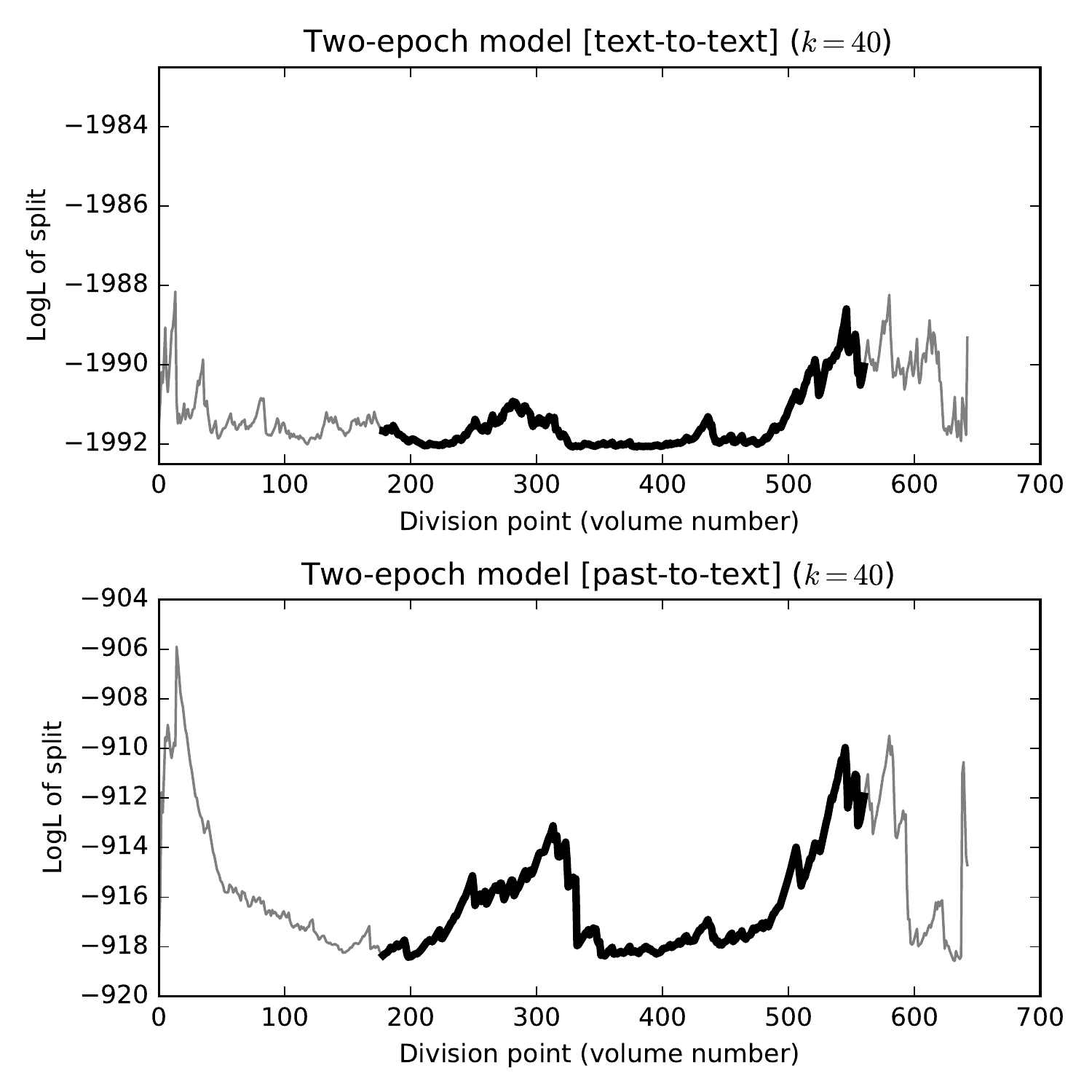}
\includegraphics[width=.31\textwidth]{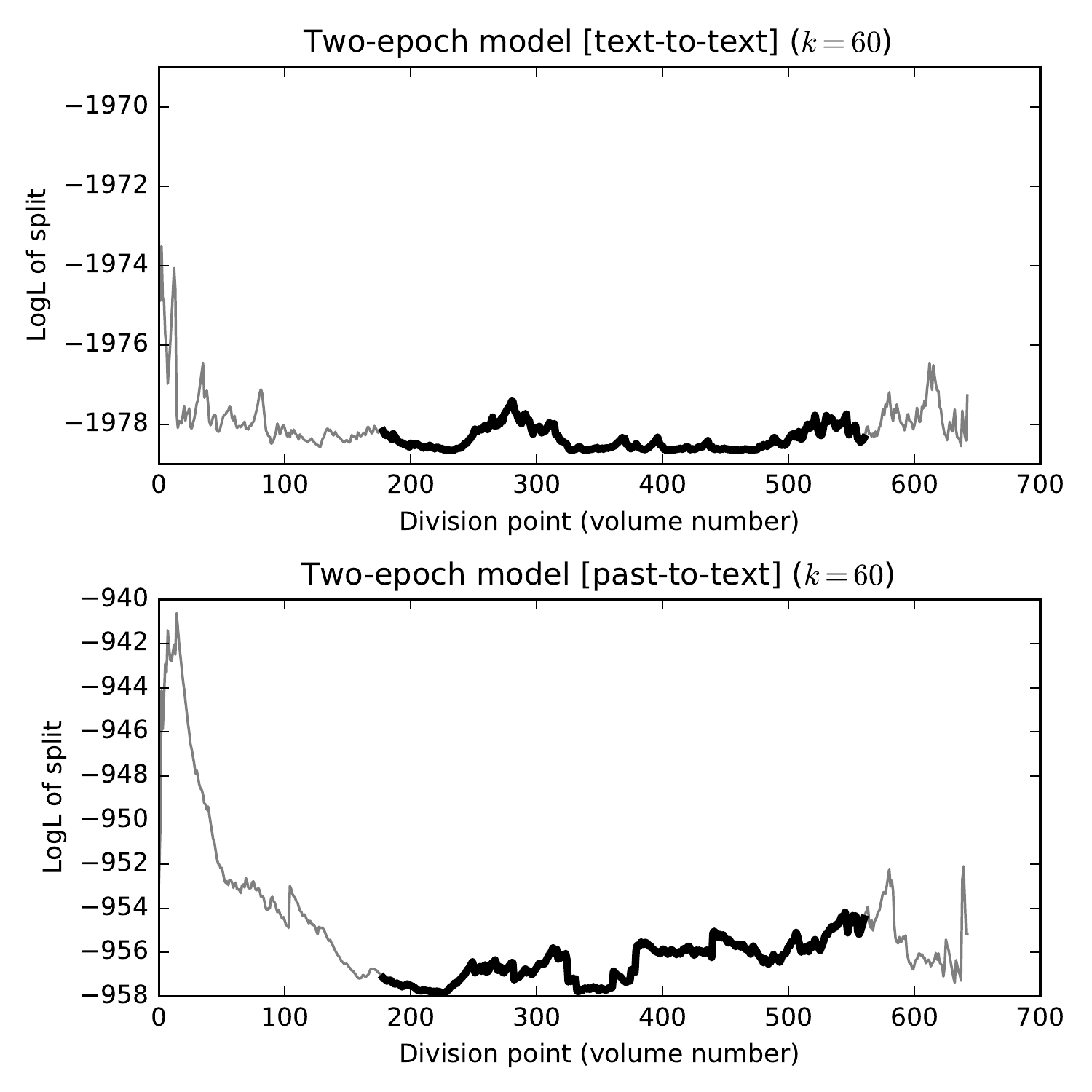}
\end{center}
\caption{\emph{Bayesian Epoch Estimation} -- Analysis of Figures \ref{fig:epoch-estimation} repeated for $k=\{20,40,60\}$.
\textbf{Left ($k=20$)} --- The text-to-text epoch break is volume 280 (1 January 1845). The text-to-past epoch break is volume 555 (27 December 1854). \textbf{Center ($k=40$)} --- The text-to-text epoch break is volume 548 (16 September 1854). The text-to-past epoch break is volume 547 (4 September 1854). Note that while a division point in 1845 is not selected in $k=40$ the text-to-past likelihood shows a local maxima at approximately volume 300. \textbf{Right ($k=60$)} --- The text-to-text epoch break is volume 282 (1 March 1845). The text-to-past epoch break is volume 547 (4 September 1854).
}
\end{figure}

\flushbottom
\clearpage
\newpage
\section*{References}
\bibliographystyle{model5-names}
\bibliography{refs.bib}

\end{document}